\documentclass[10pt]{article}
\usepackage{graphicx}
\usepackage{named}
\usepackage{amsmath}
\usepackage{amsfonts}
\usepackage{amssymb}
\usepackage{algorithm}
\usepackage{algorithmic}
\usepackage{subfigure}
\usepackage{rotating}
\usepackage{afterpage}

\def\capstyle#1{\small \emph{#1}}
\title{A Tutorial on Bayesian Optimization of Expensive Cost Functions, with Application to Active User Modeling and Hierarchical Reinforcement Learning}
\author{Eric Brochu, Vlad M. Cora and Nando de Freitas}

\newcommand{\mbs}[1]{\ensuremath{\boldsymbol{#1}}}

\newcommand{\thetav}{\mbs{\theta}}

\newcommand{\zetav}{\mbs{\zeta}}

\newcommand{\data}{{\cal D}}

\newcommand{\bv}{\mathbf{b}}
\newcommand{\cv}{\mathbf{c}}

\newcommand{\f}{\mathbf{f}}
\newcommand{\g}{\mathbf{g}}

\newcommand{\kv}{\mathbf{k}}

\newcommand{\rv}{\mathbf{r}}

\newcommand{\x}{\mathbf{x}}
\newcommand{\y}{\mathbf{y}}

\newcommand{\C}{\mathbf{C}}

\newcommand{\Hm}{\mathbf{H}}

\newcommand{\K}{\mathbf{K}}

\newcommand{\X}{\mathbf{X}}

\newcommand{\xt}{\mathbf{x}_t}

\newcommand{\EI}{\operatorname{EI}}
\newcommand{\PI}{\operatorname{PI}}
\newcommand{\LCB}{\operatorname{LCB}}
\newcommand{\UCB}{\operatorname{UCB}}
\newcommand{\GPUCB}{\operatorname{GP-UCB}}

\newcommand{\func}{(\cdot)}

\newcommand{\fmap}{\mathbf{f}_{\operatorname{MAP}}}

\newcommand{\xstar}{\mathbf{x}^{\star}}
\newcommand{\xbest}{\mathbf{x}^{+}}
\newcommand{\xlocal}{\mathbf{x}^{(\star)}}

\newcommand{\signoise}{\sigma_\mathrm{noise}}
\def\Section {\S}

\newcommand\TT{\rule{0pt}{2.6ex}}
\newcommand\BB{\rule[-1.2ex]{0pt}{0pt}}

\DeclareMathOperator*{\argmax}{argmax}
\DeclareMathOperator*{\argmin}{argmin}
\newcommand{\sfrac}[2]{\leavevmode\kern.1em
           \raise.5ex\hbox{\footnotesize #1}\kern-.1em
                   /\kern-.15em\lower.25ex\hbox{\footnotesize #2}}

\def\capstyle#1{\small \emph{#1}}

\begin{document}
\maketitle

\begin{abstract}
   We present a tutorial on Bayesian optimization, a method of finding the maximum of expensive cost functions.  Bayesian optimization employs the Bayesian technique of setting a prior over the objective function and combining it with evidence to get a posterior function.  This permits a utility-based selection of the next observation to make on the objective function, which must take into account both exploration (sampling from areas of high uncertainty) and exploitation (sampling areas likely to offer improvement over the current best observation).  We also present two detailed extensions of Bayesian optimization, with experiments---active user modelling with preferences, and hierarchical reinforcement learning---and a discussion of the pros and cons of Bayesian optimization based on our experiences.
\end{abstract}

\section{Introduction}

An enormous body of scientific literature has been devoted to the problem of optimizing a nonlinear function $f(\x)$ over a compact set $\mathcal{A}$. In the realm of optimization, this problem is formulated concisely as follows:
\[
\max_{\x \in \mathcal{A} \subset \mathbb{R}^d} f(\x)
\]
One typically assumes that the \emph{objective} function $f(\x)$ has a known mathematical representation, is convex, or is at least cheap to evaluate. Despite the influence of classical optimization on machine learning, many learning problems do not conform to these strong assumptions. Often, evaluating the objective function is expensive or even impossible, and the derivatives and convexity properties are unknown.

In many realistic sequential decision making problems, for example, one can only hope to obtain an estimate of the objective function by simulating future scenarios. Whether one adopts simple Monte Carlo simulation or adaptive schemes, as proposed in the fields of planning and reinforcement learning, the process of simulation is invariably expensive. Moreover, in some applications, drawing samples $f(\x)$ from the function corresponds to expensive processes: drug trials, destructive tests or financial investments.  In active user modeling, $\x$ represents attributes of a user query, and $f(\x)$ requires a response from the human. Computers must ask the right questions and the number of questions must be kept to a minimum so as to avoid annoying the user.

\subsection{An Introduction to Bayesian Optimization}

Bayesian optimization is a powerful strategy for finding the extrema of objective functions that are expensive to evaluate. It is applicable in situations where one does not have a closed-form expression for the objective function, but where one can obtain observations (possibly noisy) of this function at sampled values. It is particularly useful when these evaluations are costly, when one does not have access to derivatives, or when the problem at hand is non-convex. 

Bayesian optimization techniques are some of the most efficient approaches in terms of the number of function evaluations required (see, e.g. \cite{Mockus:1994,Jones:1998,Streltsov:1999,Jones:2001,Sasena:2002}).  Much of the efficiency stems from the ability of Bayesian optimization to incorporate prior belief about the problem to help direct the sampling, and to trade off exploration and exploitation of the search space.  It  is called \emph{Bayesian} because it uses the famous ``Bayes' theorem'', which states (simplifying somewhat) that the \emph{posterior} probability of a model (or theory, or hypothesis) $M$ given evidence (or data, or observations) $E$ is proportional to the \emph{likelihood} of $E$ given $M$ multiplied by the \emph{prior} probability of $M$:

\[
P(M|E) \propto P(E|M) P(M).
\]

Inside this simple equation is the key to optimizing the objective function.  In Bayesian optimization, the \emph{prior} represents our belief about the space of possible objective functions. Although the cost function is unknown, it is reasonable to assume that there exists prior knowledge about some of its properties, such as smoothness, and this makes some possible objective functions more plausible than others. 

Let's define $\x_i$ as the $i$th sample, and $f(\x_i)$ as the observation of the objective function at $\x_i$.  As we accumulate observations\footnote{
Here we use subscripts to denote sequences of data, i.e. $y_{1:t}=\{y_1,\dots,y_t\}$.
} $\data_{1:t} = \{\x_{1:t},f(\x_{1:t})\}$, the prior distribution is combined with the likelihood function $P(\data_{1:t}|f)$. Essentially, given what we think we know about the prior, how likely is the data we have seen?  If our prior belief is that the objective function is very smooth and noise-free, data with high variance or oscillations should be considered less likely than data that barely deviate from the mean.  Now, we can combine these to obtain our posterior distribution:
\[
P(f|\data_{1:t}) \propto P(\data_{1:t}|f) P(f).
\]
The posterior captures our updated beliefs about the unknown objective function.
One may also interpret this step of Bayesian optimization as estimating the objective function with a \emph{surrogate function} (also called a \emph{response surface}), described formally in \Section \ref{sec:priors} with the posterior mean function of a Gaussian process.  

\begin{figure}[t]
\begin{center}
\includegraphics[width=9cm]{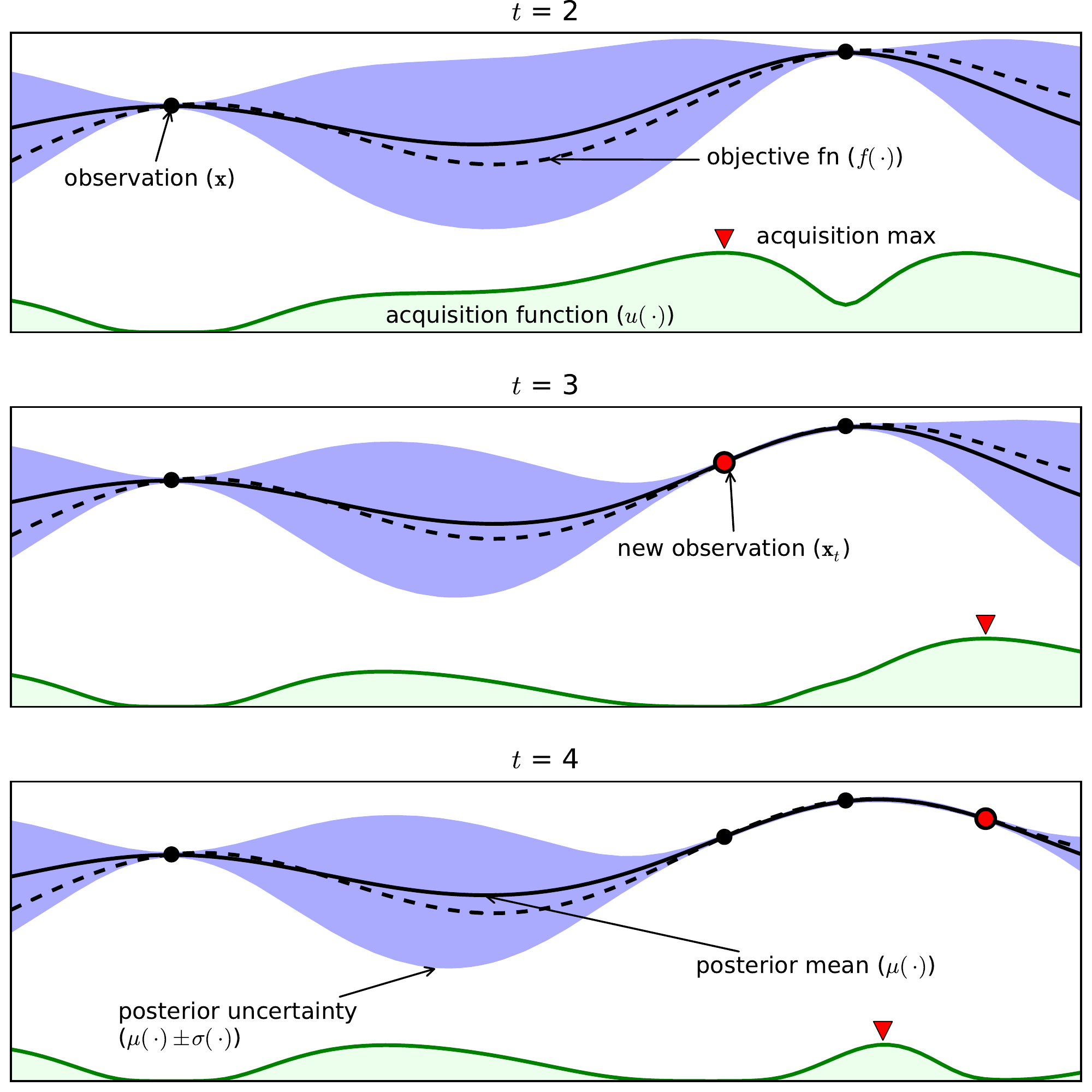}
\end{center}
\caption[Example of using Bayesian optimization on a toy 1D design problem.]{\capstyle{An example of using Bayesian optimization on a toy 1D design problem.  The figures show a Gaussian process (GP) approximation of the objective function over four iterations of sampled values of the objective function. The figure also shows the acquisition function in the lower shaded plots. The acquisition is high where the GP predicts a high objective (exploitation) and where the prediction uncertainty is high (exploration)---areas with both attributes are sampled first. Note that the area on the far left remains unsampled, as while it has high uncertainty, it is (correctly) predicted to offer little improvement over the highest observation.}}\label{fig:toyGP}
\end{figure}

To sample efficiently, Bayesian optimization uses an acquisition function to determine the next location $\x_{t+1} \in \mathcal{A}$ to sample.  The decision represents an automatic trade-off between exploration (where the objective function is very uncertain) and exploitation (trying values of $\x $ where the objective function is expected to be high).  This optimization technique has the nice property that it aims to minimize the number of objective function evaluations.  Moreover, it is likely to do well even in settings where the objective function has multiple local maxima.  

Figure~\ref{fig:toyGP} shows a typical run of Bayesian optimization on a 1D problem.    The optimization starts with two points.  At each iteration, the acquisition function is maximized to determine where next to sample from the objective function---the acquisition function takes into account the mean and variance of the predictions over the space to model the utility of sampling.  The objective is then sampled at the argmax of the acquisition function, the Gaussian process is updated and the process is repeated.
+One may also interpret this step of Bayesian optimization as estimating the objective function with a \emph{surrogate function} (also called a \emph{response surface}), described formally in \Section \ref{sec:priors} with the posterior mean function of a Gaussian process.

\subsection{Overview}

In \Section \ref{sec:bayoapproach}, we give an overview of the Bayesian optimization approach and its history.  We formally present Bayesian optimization with Gaussian process priors (\Section \ref{sec:priors}) and describe covariance functions (\Section \ref{sec:covariances}), acquisition functions (\Section \ref{sec:acquisition}) and the role of Gaussian noise (\Section \ref{sec:noise}).  In \Section \ref{sec:history}, we cover the history of Bayesian optimization, and the related fields of kriging, GP experimental design and GP active learning.

The second part of the tutorial builds on the basic Bayesian optimization model.  In \Section \ref{sec:prefgalleries} and \Section \ref{sec:hiercontrol} we discuss extensions to Bayesian optimization for active user modelling in preference galleries, and hierarchical control problems, respectively.  Finally, we end the tutorial with a brief discussion of the pros and cons of Bayesian optimization in \Section \ref{sec:discussion}.

\section{The Bayesian Optimization Approach}\label{sec:bayoapproach}

Optimization is a broad and fundamental field of mathematics.  In order to harness it to our ends, we need to narrow it down by defining the conditions we are concerned with.

Our first restriction is to simply specify that the form of the problem we are concerned with is \emph{maximization}, rather than the more common form of minimization.  The maximization of a real-valued function $\xstar = \argmax_{\x} f(\x)$ can be regarded as the minimization of the transformed function 

\[
g(\x) = -f(\x).
\]

We also assume that the objective is \emph{Lipschitz-continuous}.  That is, there exists some constant $C$, such that for all $\x_1, \x_2 \in \mathcal{A}$:
\[
\|f(\x_1)-f(\x_2)\| \leq C \|\x_1-\x_2\|,
\]
though $C$ may be (and typically is) unknown.

We can narrow the problem down further by defining it as one of \emph{global}, rather than \emph{local} optimization.  In local maximization problems, we need only find a point $\xlocal$ such that 
\[
f(\xlocal) \geq f(\x), \forall \x \textrm{ s.t. } \|\xlocal-\x\| < \epsilon.
\]
If $-f(\x)$ is convex, then any local maximum is also a global maximum.  However, in our optimization problems, we cannot assume that the negative objective function is convex.  It \emph{might} be the case, but we have no way of knowing before we begin optimizing.  

It is common in global optimization, and true for our problem, that the objective is a \emph{black box} function: we do not have an expression of the objective function that we can analyze, and   we do not know its derivatives.  Evaluating the function is restricted to querying at a point $\x$ and getting a (possibly noisy) response.  Black box optimization also typically requires that all dimensions have bounds on the search space.  In our case, we can safely make the simplifying assumption these bounds are all axis-aligned, so the search space is a hyperrectangle of dimension $d$.

A number of approaches exist for this kind of global optimization and have been well-studied in the literature (e.g.,\ \cite{Torn:1989,Mongeau:1998,Liberti:2006,Zhigljavsky:2008}).  Deterministic approaches include interval optimization and branch and bound methods.
\emph{Stochastic approximation} is a popular idea for optimizing unknown objective functions in machine learning contexts~\cite{Kushner:1997}. It is the core idea in most reinforcement learning algorithms \cite{Bertsekas:1996,Sutton:1998}, learning methods for Boltzmann machines and deep belief networks \cite{Younes:1989,Hinton:2006} and parameter estimation for nonlinear state space models \cite{Poyiadjis:2005,Martinez--Cantin:2006}. However, these are generally unsuitable for our domain because they still require many samples, and in the active user-modelling domain drawing samples is \emph{expensive}.

Even in a noise-free domain, evaluating an objective function with Lipschitz continuity $C$ on a $d$-dimensional unit hypercube, guaranteeing the best observation $f(\x^+) \geq f(\xstar) - \epsilon$ requires $(C/2\epsilon)^d$ samples \cite{Betro:1991}.  This can be an incredibly expensive premium to pay for insurance against unlikely scenarios.  As a result, the idea naturally arises to relax the guarantees against pathological worst-case scenarios.  The goal, instead, is to use evidence and prior knowledge to maximize the posterior at each step, so that each new evaluation decreases the distance between the true global maximum and the expected maximum given the model.  This is sometimes called ``one-step'' \cite{Mockus:1994} ``average-case'' \cite{Streltsov:1999} or ``practical'' \cite{Lizotte:2008} optimization.  This average-case approach has weaker demands on computation than the worst-case approach. As a result, it may provide faster solutions in many practical domains where one does not believe the worst-case scenario is plausible.  

Bayesian optimization uses the prior and evidence to define a posterior distribution over the space of functions.  The Bayesian model allows for an elegant means by which informative priors can describe attributes of the objective function, such as smoothness or the most likely locations of the maximum, even when the function itself is not known.  Optimizing follows the principle of \emph{maximum expected utility}, or, equivalently, \emph{minimum expected risk}.  The process of deciding where to sample next requires the choice of a utility function and a way of optimizing the expectation of this utility with respect to the posterior distribution of the objective function. This secondary optimization problem is usually easier because the utility is typically chosen so that it is easy to evaluate, though still nonconvex.  To make clear which function we are discussing, we will refer to this utility as the \emph{acquisition function} (also sometimes called the \emph{infill function}).  In \Section \ref{sec:acquisition}, we will discuss some common acquisition functions.

In practice, there is also have the possibility of measurement noise, which we will assume is Gaussian.  We define $\x_i$ as the $i$th sample and $y_i = f(\x_i) + \varepsilon_i,$ with $\varepsilon_i \stackrel{iid}{\sim} \mathcal{N}(0, \signoise^2)$, as the noisy observation of the objective function at $\x_i$. We will discuss noise in more detail in \Section \ref{sec:noise}.

The Bayesian optimization procedure is shown in Algorithm~\ref{alg:bayopt}. As mentioned earlier, it has two components: the posterior distribution over the objective and the acquisition function. Let us focus on the posterior distribution first and come back to the acquisition function in \Section \ref{sec:acquisition}.
As we accumulate observations $\data_{1:t} = \{\x_{1:t},y_{1:t}\}$, a prior distribution $P(f)$ is combined with the likelihood function $P(\data_{1:t}|f)$ to produce the posterior distribution:
$
P(f|\data_{1:t}) \propto P(\data_{1:t}|f) P(f).
$
The posterior captures the updated beliefs about the unknown objective function.
One may also interpret this step of Bayesian optimization as estimating the objective function with a \emph{surrogate function} (also called a \emph{response surface}).  In \Section \ref{sec:priors}, we will discuss how Gaussian process priors can be placed on $f$.

\begin{algorithm}[t!]
\caption{Bayesian Optimization}\label{alg:bayopt}
\begin{algorithmic}[1]
{\footnotesize
   \FOR{$t=1,2,\dots$}
       \STATE Find $\x_t$ by optimizing the acquisition function over the GP: $\x_t = \argmax_{\x}u(\x|\data_{1:t-1})$.
       \STATE Sample the objective function: $y_t=f(\x_t)+\varepsilon_t$.
       \STATE Augment the data $\data_{1:t} = \{\data_{1:t-1}, (\x_t, y_t)\}$ and update the GP.
   \ENDFOR
}
\end{algorithmic}
\end{algorithm}


\subsection{Priors over functions}\label{sec:priors}

Any Bayesian method depends on a prior distribution, by definition.  A Bayesian optimization method will converge to the optimum if (i) the acquisition function is continuous and approximately minimizes the risk (defined as the expected deviation from the global minimum at a fixed point $\x$); and (ii) conditional variance converges to zero (or appropriate positive minimum value in the presence of noise) if and only if the distance to the nearest observation is zero \cite{Mockus:1982,Mockus:1994}.  Many models could be used for this prior---early work mostly used the Wiener process (\Section \ref{sec:history}).  However, Gaussian process (GP) priors for Bayesian optimization date back at least to the late 1970s \cite{OHagan:1978,Zilinskas:1980}.  Mo\v{c}kus \shortcite{Mockus:1994} explicitly set the framework for the Gaussian process prior by specifying the additional ``simple and natural'' conditions that (iii) the objective is continuous; (iv) the prior is homogeneous; (v) the optimization is independent of the $m^{\mathrm{th}}$ differences.  This includes a very large family of common optimization tasks, and Mo\v{c}kus showed that the GP prior is well-suited to the task.

\begin{figure}[t]
 \begin{center}
   \includegraphics[width=\textwidth]{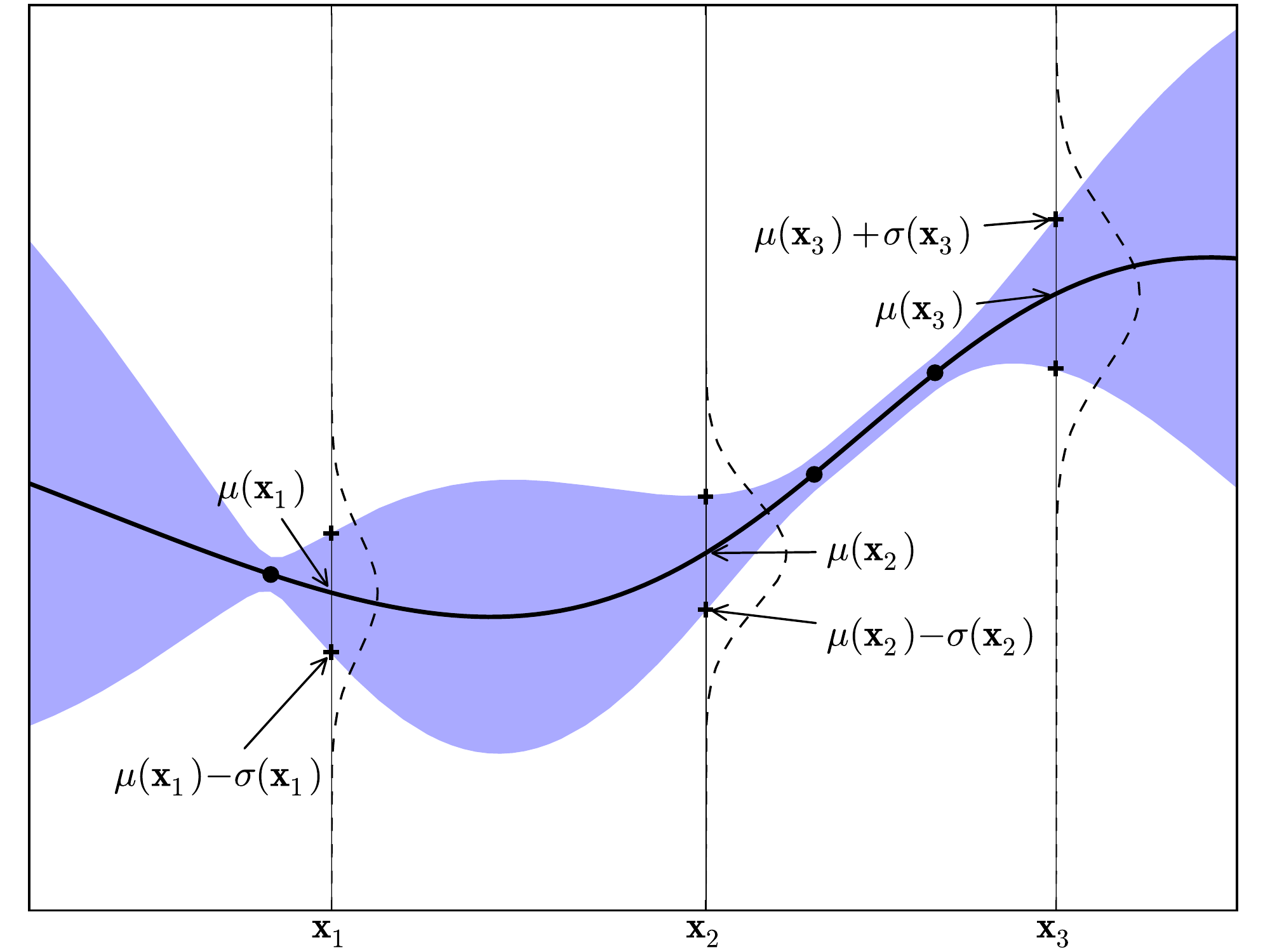}
 \end{center}
 \caption[Simple 1D Gaussian process with three observations.]{\capstyle{Simple 1D Gaussian process with three observations.  The solid black line is the GP surrogate mean prediction of the objective function given the data, and the shaded area shows the mean plus and minus the variance.  The superimposed Gaussians correspond to the GP mean and standard deviation ($\mu\func$ and $\sigma\func$) of prediction at the points, $\x_{1:3}$.}}\label{fig:GPvar}
\end{figure}

A GP is an extension of the multivariate Gaussian distribution to an infinite-dimension stochastic process for which any finite combination of dimensions will be a Gaussian distribution.  Just as a Gaussian distribution is a distribution over a random variable, completely specified by its mean and covariance, a GP is a distribution over functions, completely specified by its mean function, $m$ and covariance function, $k$:
\begin{equation*}
f(\x) \sim \mathcal{GP}(m(\x), k(\x, \x')).
\end{equation*}

It is often useful to intuitively think of a GP as analogous to a function, but instead of returning a scalar $f(\x)$ for an arbitrary $\x$, it returns the mean and variance of a normal distribution (Figure \ref{fig:GPvar}) over the possible values of $f$ at $\x$. Stochastic processes are sometimes called ``random functions'', by analogy to random variables. 

For convenience, we assume here that the prior mean is the zero function $m(\x) = 0$; alternative priors for the mean can be found in, for example \cite{Martinez--Cantin:2009,Brochu:2010}.  This leaves us the more interesting question of defining the covariance function $k$.  A very popular choice is the squared exponential function:
\begin{equation}
    k(\x_i, \x_j) = \exp\left(- \frac{1}{2}\left\|\x_i-\x_j\right\|^2 \right).\label{eqn:sqexp}
\end{equation}
Note that this function approaches 1 as values get close together and 0 as they get further apart.  Two points that are close together can be expected to have a very large influence on each other, whereas distant points have almost none.  This is a necessary condition for convergence under the assumptions of \cite{Mockus:1994}.  We will discuss more sophisticated kernels in \Section \ref{sec:covariances}.

If we were to sample from the prior, we would choose $\{\x_{1:t}\}$ and sample the values of the function at these indices to 
produce the pairs $\{\x_{1:t},\f_{1:t}\}$, where $\f_{1:t}=f(\x_{1:t})$. The function values are drawn according to a multivariate normal distribution $\mathcal{N}(0,\K)$, where the kernel matrix is given by:
\begin{eqnarray*}
    \K = \left[\begin{matrix}
            k(\x_1, \x_1) & \ldots & k(\x_1, \x_t)\\
            \vdots & \ddots & \vdots\\
            k(\x_t, \x_1) & \ldots & k(\x_t, \x_t)\\
        \end{matrix}\right].
\end{eqnarray*}
Of course, the diagonal values of this matrix are 1 (each point is perfectly correlated with itself), which is only possible in a noise-free environment.  We will discuss noise in \Section \ref{sec:noise}. Also, recall that we have for simplicity chosen the zero mean function.

In our optimization tasks, however, we will use data from an external model to fit the GP and get the posterior.  Assume that we already have the observations $\{\x_{1:t},\f_{1:t}\}$, say from previous iterations, and that we want to use Bayesian optimization to decide what point $\x_{t+1}$ should be considered next. Let us denote the value of the function at this arbitrary point as $f_{t+1}=f(\x_{t+1})$. Then, by the properties of Gaussian processes, $\f_{1:t}$ and $f_{t+1}$ are jointly Gaussian:
\[
\begin{bmatrix}
    \f_{1:t} \\
    f_{t+1}
\end{bmatrix}
\sim {\cal N} 
\left( \mathbf{0} ,
    \begin{bmatrix}
        \K & \kv \\
        \kv^{T} & k(\x_{t+1},\x_{t+1})
    \end{bmatrix}
\right),
\]
where
\[
\kv =
\begin{bmatrix}
    k(\x_{t+1},\x_1) & k(\x_{t+1},\x_2) & \cdots & k(\x_{t+1},\x_t) \\
\end{bmatrix}
\]
Using the Sherman-Morrison-Woodbury formula (see, e.g., \cite{Rasmussen:2006,Press:2007}), one can easily arrive 
at an expression for the predictive distribution:
\[
P(f_{t+1}|\data_{1:t}, \x_{t+1}) = {\cal N} \left(\mu_t(\x_{t+1}), \sigma^2_t(\x_{t+1})\right)
\]
where
\begin{eqnarray*}
\mu_t(\x_{t+1})&=&
\mathbf{k}^T \mathbf{K}^{-1} \f_{1:t} \\
\sigma_t^2(\x_{t+1})&=& k(\x_{t+1},\x_{t+1}) - \mathbf{k}^T \mathbf{K}^{-1}\mathbf{k}.
\nonumber
\end{eqnarray*}
That is, $\mu_t\func$ and $\sigma_t^2\func$ are the sufficient statistics of the predictive posterior distribution $P(f_{t+1}|\data_{1:t}, \x_{t+1})$.  For legibility, we will omit the subscripts on $\mu$ and $\sigma$ except where it might be unclear.  In the sequential decision making setting, the number of query points is relatively small and, consequently, the GP predictions are easy to compute.


\begin{figure}[t]
 \begin{center}
   \includegraphics[width=\textwidth]{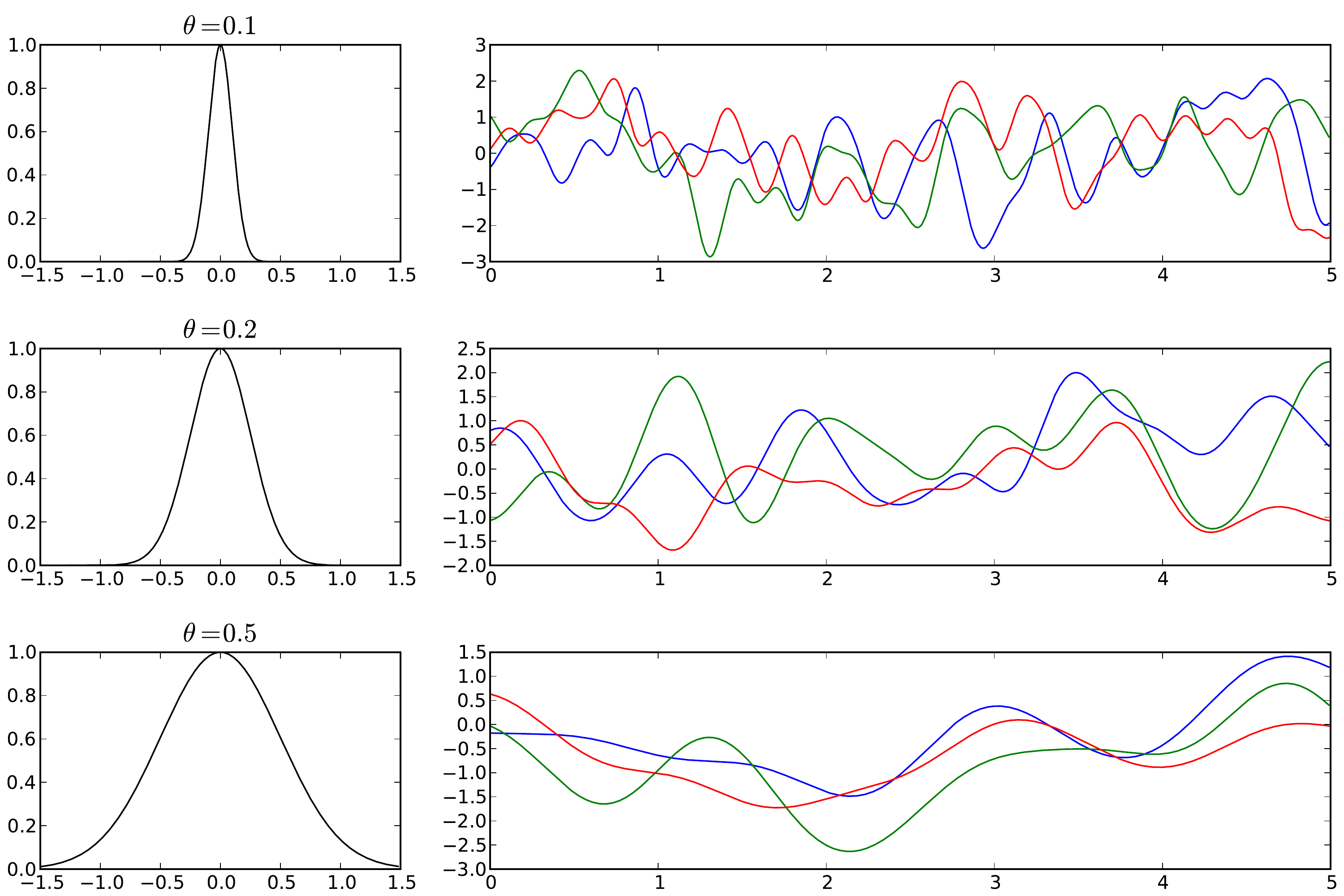}
 \end{center}
 \caption[The effect of changing the kernel hyperparameters.]{\capstyle{The effect of changing the kernel hyperparameters.  Shown are squared exponential kernels with $\theta = 0.1, 0.2, 0.5$.  On the left is the function $k(0, \x)$.  On the right are some one-dimensional functions sampled from a GP with the hyperparameter value.}}\label{fig:kernels}
\end{figure}

\subsection{Choice of covariance functions}\label{sec:covariances}

The choice of covariance function for the Gaussian Process is crucial, as it determines the smoothness properties of samples drawn from it.  The squared exponential kernel in Eqn~(\ref{eqn:sqexp}) is actually a little naive, in that divergences of all features of $\x$ affect the covariance equally.  

Typically, it is necessary to generalize by adding \emph{hyperparameters}.  In an isotropic model, this can be done with a single hyperparameter $\theta$, which controls the width of the kernel:
\begin{equation*}
    k(\x_i, \x_j) = \exp\left(- \frac{1}{2\theta^2}\left\|\x_i-\x_j\right\|^2 \right).
\end{equation*}
For anisotropic models, a very popular choice is the squared exponential kernel with a vector of automatic relevance determination (ARD) hyperparameters $\thetav$ \cite[page 106]{Rasmussen:2006}:
\[
k(\x_i, \x_j) = \exp\big(-\tfrac{1}{2}(\x_i-\x_j)^{T}
\operatorname{diag}(\thetav)^{-2}(\x-\x')\big), 
\]
where $\operatorname{diag}(\thetav)$ is a diagonal matrix with $d$ entries $\thetav$ along the diagonal. Intuitively, if a particular $\theta_\ell$ has a small value, the kernel becomes independent of $\ell$-th input, effectively removing it automatically. Hence, irrelevant dimensions are discarded.  Figure~\ref{fig:kernels} shows examples of different hyperparameter values on the squared exponential function and what functions sampled from those values look like.  Typically, the hyperparameter values are learned by ``seeding'' with a few random samples and maximizing the log-likelihood of the evidence given $\thetav$ \cite{Jones:1998,Sasena:2002,Santner:2003,Rasmussen:2006}.  This can often be aided with an informative hyperprior on the hyperparameters, often a log normal prior \cite{Lizotte:2008,Frean:2008}.  Methods of learning these values more efficiently is currently an active subfield of research (\emph{e.g.}~\cite{Osborne:2010,Brochu:2010}).

Another important kernel for Bayesian optimization is the Mat\'{e}rn kernel \cite{Matern:1960,Stein:1999}, which incorporates a smoothness parameter $\varsigma$ to permit greater flexibility in modelling functions:

\begin{equation*}
    k(\x_i, \x_j) = \frac{1}{2^{\varsigma-1}\Gamma(\varsigma)}\left(2 \sqrt{\varsigma}\left\|\x_i-\x_j\right\|\right)^\varsigma H_{\varsigma} \left(2\sqrt{\varsigma} \left\|\x_i-\x_j\right\|\right),
\end{equation*}
where $\Gamma\func$ and $H_\varsigma\func$ are the Gamma function and the Bessel function of order $\varsigma$.  Note that as $\varsigma \rightarrow \infty$, the Mat\'{e}rn kernel reduces to the squared exponential kernel, and when $\varsigma = 0.5$, it reduces to the unsquared exponential kernel.  As with the squared exponential, length-scale hyperparameter are often incorporated.

While the squared exponential and Mat\'{e}rn are the most common kernels for GPs, numerous others have been examined in the machine learning literature (see, e.g., \cite{Genton:2001} or \cite[Chapter~4]{Rasmussen:2006} for an overview).  Appropriate covariance functions can also be used to extend the model in other interesting ways.  For example, the recent sequential sensor work of Osborne, Garnett and colleagues uses GP models with extensions to the covariance function to model the characteristics of changepoints \cite{Osborne:2010a} and the locations of sensors in a network \cite{Garnett:2010}.  A common additional hyperparameter is simply a scalar applied to $k$ to control the magnitude of the variance.  

Determining which of a set of possible kernel functions to use for a problem typically requires a combination of engineering and automatic model selection, either hierarchical Bayesian model selection~\cite{Mackay:1992} or cross-validation.  However, these methods require fitting a model given a representative sample of data.  In \cite{Brochu:2010}, we discuss how model selection can be performed using models believed to be similar.  The techniques introduced in \cite{Brochu:2010a} could also be applied to model selection, though that is outside the scope of this tutorial.

\subsection{Acquisition Functions for Bayesian Optimization}\label{sec:acquisition}

Now that we have discussed placing priors over smooth functions and how to update these priors in light of new observations, we will focus our attention on the acquisition component of Bayesian optimization.  The role of the acquisition function is to guide the search for the optimum.  Typically, acquisition functions are defined such that high acquisition corresponds to \emph{potentially} high values of the objective function, whether because the prediction is high, the uncertainty is great, or both.  Maximizing the acquisition function is used to select the next point at which to evaluate the function.  That is, we wish to sample $f$ at $\argmax_\x u(\x|\data)$, where $u\func$ is the generic symbol for an acquisition function.

\subsubsection{Improvement-based acquisition functions}\label{sec:piei}

The early work of Kushner \shortcite{Kushner:1964} suggested maximizing the \emph{probability of improvement} over the incumbent $f(\xbest)$, where $\x^+ = \argmax_{\x_i \in \x_{1:t}} f(\x_i)$, so that
\begin{eqnarray*}
\PI(\x) &=& P(f(\x) \geq f(\xbest))\\
 &=& \Phi\left(\frac{\mu(\x) - f(\xbest)}{\sigma(\x)}\right),
\end{eqnarray*}
where $\Phi\func$ is the normal cumulative distribution function.  This function is also sometimes called \emph{MPI} (for ``maximum probability of improvement'') or ``the $P$-algorithm'' (since the utility is the probability of improvement).

The drawback, intuitively, is that this formulation is pure exploitation.  Points that have a high probability of being infinitesimally greater than $f(\xbest)$ will be drawn over points that offer larger gains but less certainty.  As a result, a modification is to add a trade-off parameter $\xi \geq 0$:
\begin{eqnarray}
\PI(\x) &=& P(f(\x) \geq f(\xbest)+\xi)\nonumber\\
 &=& \Phi\left(\frac{\mu(\x) - f(\xbest) - \xi}{\sigma(\x)}\right),\label{eqn:pi}
\end{eqnarray}
The exact choice of $\xi$ is left to the user, though Kushner recommended a schedule for $\xi$, so that it started fairly high early in the optimization, to drive exploration, and decreased toward zero as the algorithm continued.   Several researchers have studied the empirical impact of different values of $\xi$ in different domains \cite{Torn:1989,Jones:2001,Lizotte:2008}.  

An appealing characteristic of this formulation for perceptual and preference models is that while maximizing $\PI\func$ is still greedy, it selects the point most likely to offer an improvement of at least $\xi$.  This can be useful in psychoperceptual tasks, where there is a threshold of perceptual difference.

\begin{figure}[t]
 \begin{center}
   \includegraphics[width=\textwidth]{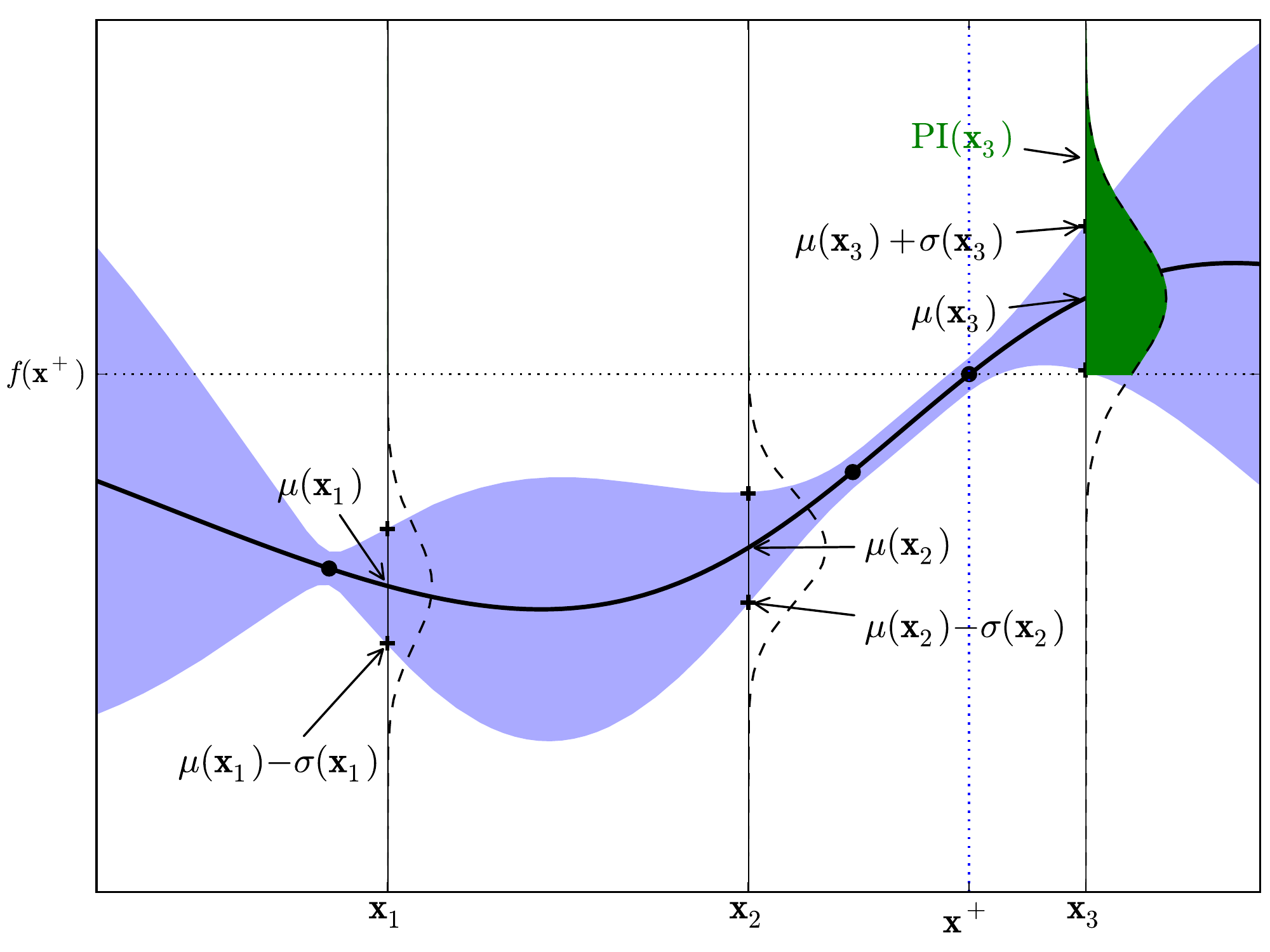}
 \end{center}
 \caption[Gaussian process from Figure~\ref{fig:GPvar}, additionally showing the region of probable improvement.]{\capstyle{Gaussian process from Figure~\ref{fig:GPvar}, additionally showing the region of probable improvement.  The maximum observation is at $\xbest$. The darkly-shaded area in the superimposed Gaussian above the dashed line can be used as a measure of improvement, $I(\x)$.  The model predicts almost no possibility of improvement by observing at $\x_1$ or $\x_2$, while sampling at $\x_3$ is more likely to improve on $f(\xbest)$.}}
 \label{fig:GPvarEI}
\end{figure}

Jones \shortcite{Jones:2001} notes that the performance of $\PI\func$
\begin{quote}
``is truly impressive.  It would be quite natural if the reader, like so many others, became enthusiastic about this approach.  But if there is a single lesson to be taken away from this paper, it is that nothing in this response-surface area is so simple.  There always seems to be a counterexample.  In this case, the difficulty is that [the $\PI\func$ method] is extremely sensitive to the choice of the target.  If the desired improvement is too small, the search will be highly local and will only move on to search globally after searching nearly exhaustively around the current best point.  On the other hand, if [$\xi$] is set too high, the search will be excessively global, and the algorithm will be slow to fine-tune any promising solutions.''
\end{quote}

A somewhat more satisfying alternative acquisition function would be one that takes into account not only the probability of improvement, but the magnitude of the improvement a point can potentially yield.  In particular, we want to minimize the expected deviation from the true maximum $f(\xstar)$, when choosing a new trial point:
\begin{eqnarray}
\x_{t+1} &=&  \argmin_{\x} \Bbb{E}(\|f_{t+1}(\x) - f(\xstar)\| \;|\data_{1:t}) \nonumber \\
&=& \argmin_{\x} \int \|f_{t+1}(\x) - f(\xstar)\| P(f_{t+1}|\data_{1:t})df_{t+1},
\nonumber 
\end{eqnarray}
Note that this decision process is myopic in that it only considers one-step-ahead choices. However, if we want to plan two steps ahead, we can easily apply recursion:
\[
\x_{t+1} = \argmin_{\x}\Bbb{E}\left(\min_{\x'} \Bbb{E}(\|f_{t+2}(\x')
 - f(\xstar)\|    \;| \data_{t+1}) \; | \data_{1:t}\right)
\]
One could continue applying this procedure of dynamic programming for as many steps ahead as desired. However, because of its expense,  Mo\v{c}kus \emph{et al.}\ \shortcite{Mockus:1978} proposed the alternative of maximizing the expected improvement with respect to $f(\xbest)$. Specifically, Mo\v{c}kus defined the improvement function as:
\[
\operatorname{I}(\x) = \max\{0, f_{t+1}(\x) - f(\xbest)\}.
\]
That is, $\operatorname{I}(\x)$ is positive when the prediction is higher than the best value known thus far.  Otherwise, $\operatorname{I}(\x)$ is set to zero. The new query point is found by maximizing the expected improvement:
\[
\x =  \argmax_{\x} \Bbb{E}(\max\{0,f_{t+1}(\x) - f(\xbest)\} \;|\data_t)
\]
The likelihood of improvement $\operatorname{I}$ on a normal posterior distribution characterized by $\mu(\x), \sigma^2(\x)$ can be computed from the normal density function,
\[
\frac{1}{\sqrt{2\pi}\sigma(\x)} \exp\left(-\frac{(\mu(\x) - f(\xbest)-\operatorname{I})^2}{2\sigma^2(\x)}\right).
\]
The expected improvement is the integral over this function:
\begin{eqnarray*}
\Bbb{E}(\operatorname{I}) &=& \int_{\operatorname{I}=0}^{\operatorname{I}=\infty}\operatorname{I}\frac{1}{\sqrt{2\pi}\sigma(\x)} \exp\left(-\frac{(\mu(\x) - f(\xbest)-\operatorname{I})^2}{2\sigma^2(\x)}\right) d\operatorname{I}\\
&=& \sigma(\x)\left[ \frac{\mu(\x)-f(\xbest)}{\sigma(\x)}\Phi\left(\frac{\mu(\x)-f(\xbest)}{\sigma(\x)}\right)+\phi\left(\frac{\mu(\x)-f(\xbest)}{\sigma(\x)}\right) \right]
\end{eqnarray*}

The expected improvement can be evaluated analytically \cite{Mockus:1978,Jones:1998}, yielding:
\begin{eqnarray}\label{eqn:EGO_EI}
  \EI(\x) &=& \left\{ \begin{array}{ll}
      (\mu(\x)-f(\xbest))\Phi(Z) +\sigma(\x)
\phi(Z) & \textrm{if } \sigma(\x) > 0\\
      0 & \textrm{if } \sigma(\x) = 0
  \end{array} \right. \\
  Z &=& \frac{\mu(\x) - f(\xbest)}{\sigma(\x)}\nonumber
\end{eqnarray}
where $\phi\func$ 
and $\Phi\func$ denote the PDF and CDF of the standard normal distribution respectively.  Figure~\ref{fig:GPvarEI} illustrates a typical expected improvement scenario.

It should be said that being myopic is not a requirement here. For example, it is possible to derive analytical expressions for the two-step ahead expected improvement \cite{Ginsbourger:2008} and multistep Bayesian optimization \cite{Garnett:2010a}. This is indeed a very promising recent direction.

\subsubsection{Exploration-exploitation trade-off}\label{sec:explparam}

The expectation of the improvement function with respect to the predictive distribution of the Gaussian process enables us to balance the trade-off of exploiting and exploring. When exploring, we should choose points where the surrogate variance is large. When exploiting, we should choose points where the surrogate mean is high.

It is highly desirable for our purposes to express $\EI\func$ in a generalized form which controls the trade-off between global search and local optimization (exploration/exploitation).  Lizotte \shortcite{Lizotte:2008} suggests a $\xi \geq 0$ parameter such that:
\begin{eqnarray}
\EI(\x) &=& 
   \left\{ \begin{array}{ll}
          (\mu(\x) - f(\xbest) - \xi)\Phi(Z) +\sigma(\x)
   \phi(Z) & \textrm{if } \sigma(\x) > 0\\
          0 & \textrm{if } \sigma(\x) = 0
      \end{array} \right. ,\label{eqn:ei}
\end{eqnarray}
where
\begin{eqnarray*}
Z &=& 
   \left\{ \begin{array}{ll}
   \frac{\mu(\x) - f(\xbest) - \xi}{\sigma(\x)} & \textrm{if } \sigma(\x) > 0\\
          0 & \textrm{if } \sigma(\x) = 0
      \end{array} \right. .
\end{eqnarray*}
This $\xi$ is very similar in flavour to the $\xi$ used in Eqn~(\ref{eqn:pi}), and to the approach used by Jones \emph{et al.} \shortcite{Jones:2001}.  Lizotte's experiments suggest that setting $\xi= 0.01$ (scaled by the signal variance if necessary) works well in almost all cases, and interestingly, setting a cooling schedule for $\xi$ to encourage exploration early and exploitation later does \emph{not} work well empirically, contrary to intuition (though Lizotte did find that a cooling schedule for $\xi$ might slightly improve performance on short runs ($t<30$) of PI optimization).

\begin{figure}[t]
 \begin{center}
   \includegraphics[width=\textwidth]{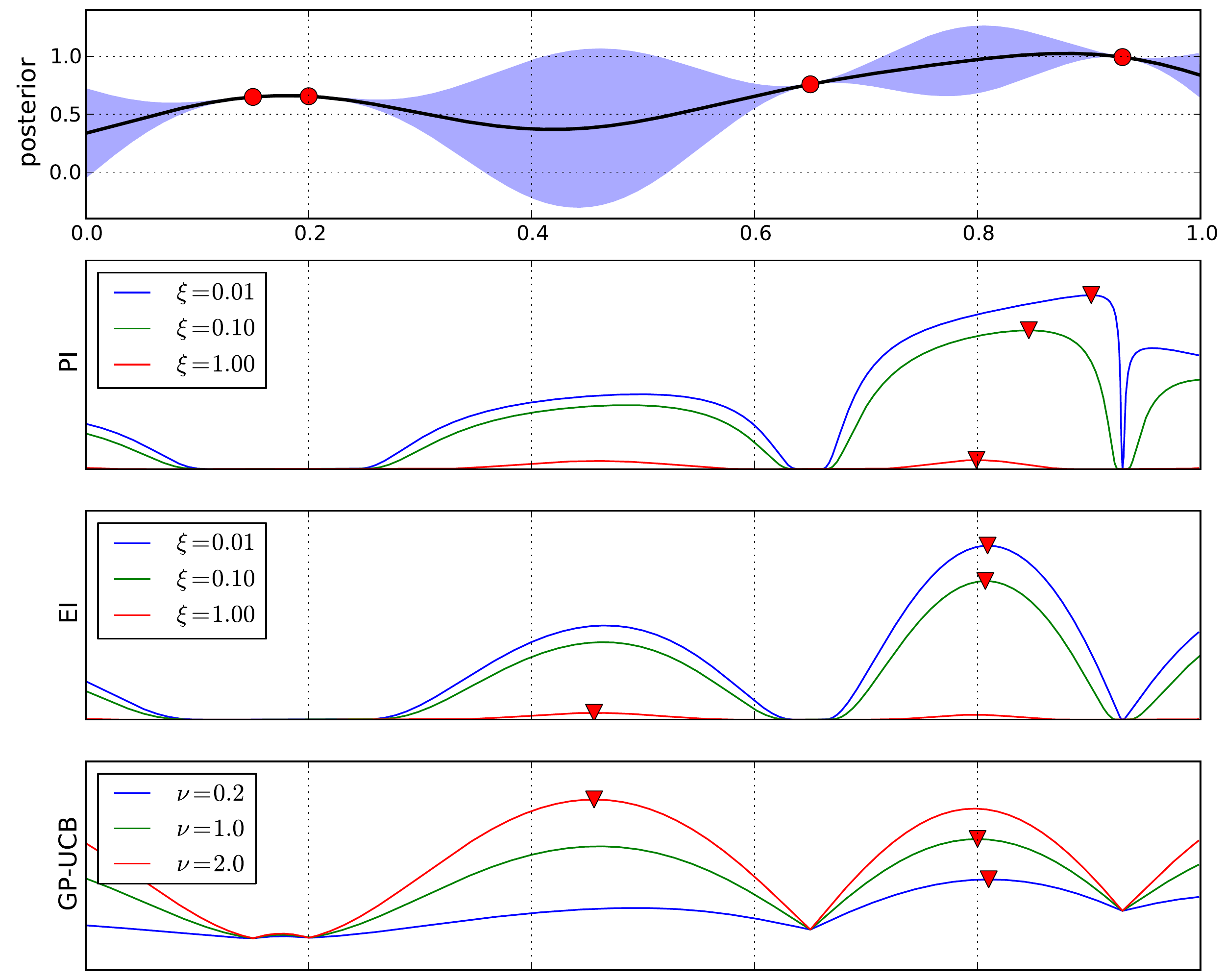}
 \end{center}
 \caption[Examples of acquisition functions and their settings.]{\capstyle{Examples of acquisition functions and their settings.  The GP posterior is shown at top.  The other images show the acquisition functions for that GP.  From the top: probability of improvement (Eqn~(\ref{eqn:pi})), expected improvement (Eqn~(\ref{eqn:ei})) and upper confidence bound (Eqn~(\ref{eqn:ucb})).  The maximum of each function is shown with a triangle marker.}} 
 \label{fig:acquisition}
\end{figure}

\begin{figure}[ht]
 \begin{center}
   \includegraphics[width=\textwidth]{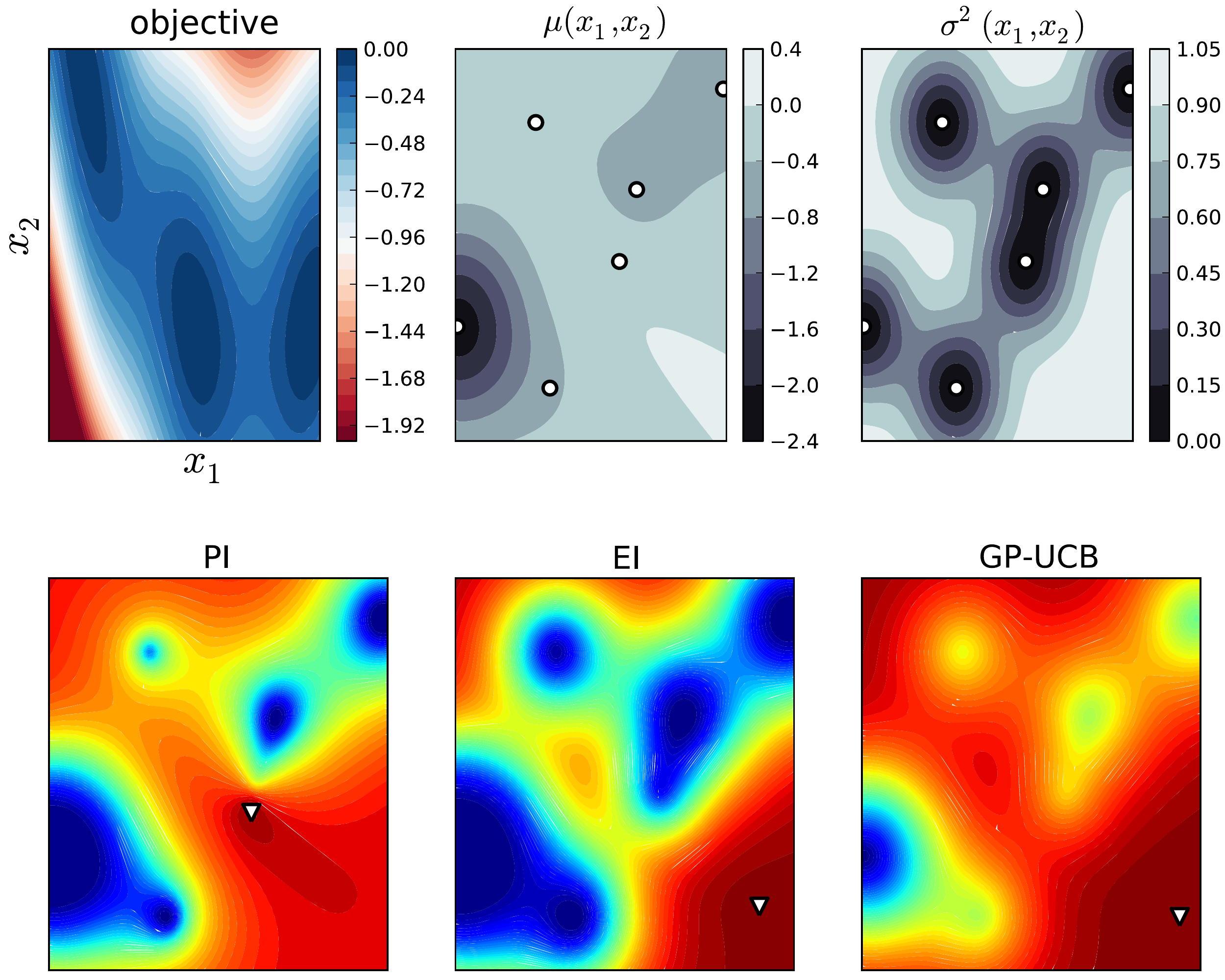}
 \end{center}
 \caption[Examples of acquisition functions and their settings in 2 dimensions.]{\capstyle{Examples of acquisition functions and their settings in 2 dimensions.  The top row shows the objective function (which is the Branin function here), and the posterior mean and variance estimates $\mu\func$ and $\sigma^2\func$.  The samples used to train the GP are shows with white dots.  The second row shows the acquisition functions for the GP.  From left to right: probability of improvement (Eqn~(\ref{eqn:pi})), expected improvement (Eqn~(\ref{eqn:ei})) and upper confidence bound (Eqn~(\ref{eqn:ucb})).  The maximum of each function is shown with a triangle marker.}}
 \label{fig:acquisition2d} 
\end{figure}

\subsubsection{Confidence bound criteria}\label{sec:ucb}


\begin{sidewaysfigure}[htp!]
 \begin{center}
   \includegraphics[width=17cm]{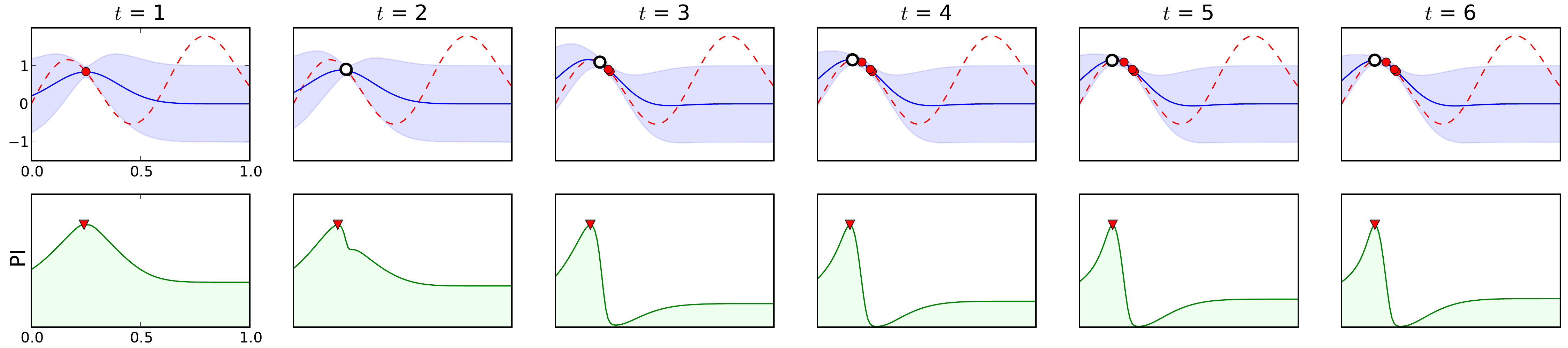}\\
   \vspace{.5cm}
   \includegraphics[width=17cm]{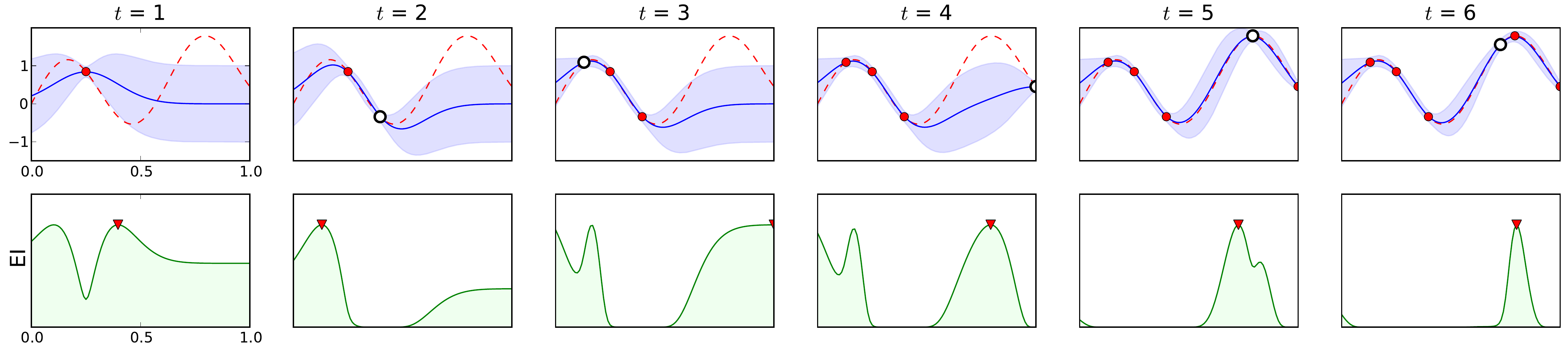}\\
   \vspace{.5cm}
   \includegraphics[width=17cm]{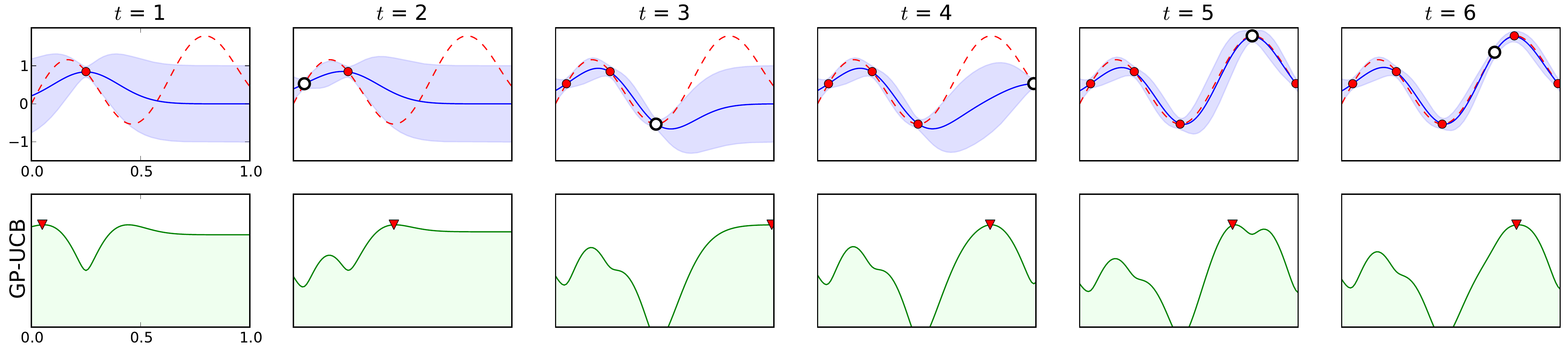}
 \end{center}
 \caption[Comparison of acquisition functions on a toy 1D problem.]{\capstyle{Comparison of probability of improvement (top), expected improvement (middle) and upper confidence bound (bottom) acquisition functions on a toy 1D problem.  In the upper rows, the objective function is shown with the dotted red line, the solid blue line is the GP posterior mean.  In the lower rows, the respective infill functions are shown, with a star denoting the maximum.  The optimizations are initialized with the same two points, but quickly follow different sampling trajectories.  In particular, note that the greedy $\EI$ algorithm ignores the region around $x=0.4$ once it is determined there is minimal chance of improvement, while $\GPUCB$ continues to explore.}}
 \label{fig:eivsucb} 
\end{sidewaysfigure}

Cox and John \shortcite{Cox:1992,Cox:1997} introduce an algorithm they call ``Sequential Design for Optimization'', or \emph{SDO}.  Given a random function model, SDO selects points for evaluation based on the \emph{lower confidence bound} of the prediction site:
\[
\LCB(\x) = \mu(\x) - \kappa \sigma(x),
\]
where $\kappa \geq 0$.  While they are concerned with minimization, we can maximize by instead defining the upper confidence bound:
\[
\UCB(\x) = \mu(\x) + \kappa \sigma(x).
\]

Like other parameterized acquisition models we have seen, the parameter $\kappa$ is left to the user.  However, an alternative acquisition function has been proposed by Srinivas \emph{et al.}\ \shortcite{Srinivas:2010}.  Casting the Bayesian optimization problem as a multi-armed bandit, the acquisition is the instantaneous regret function 
\[
r(\x) = f(\xstar)-f(\x).
\]
The goal of optimizing in the framework is to find:
\[
\min \sum_t^T r(\xt) = \max \sum_t^T f(\x_t),
\]
where $T$ is the number of iterations the optimization is to be run for.

Using the \emph{upper confidence bound} selection criterion with $\kappa_t=\sqrt{\nu\tau_t}$ and the hyperparameter $\nu>0$ Srinivas \emph{et al.}\ define
\begin{equation}
\operatorname{GP-UCB}(\x) = \mu(\x) + \sqrt{\nu\tau_t}\sigma(\x).\label{eqn:ucb}
\end{equation}
With $\nu=1$ and $\tau_t = 2 \log(t^{d/2+2}\pi^2/3\delta)$, it can be shown\footnote
{
These bounds hold for \emph{reasonably smooth} kernel functions, where the exact formulation of the bounds depends upon the form of kernel used. We refer the interested reader to the original paper \cite{Srinivas:2010}.
}
with high probability that this method is \emph{no regret}, i.e.\  $\lim_{T\to\infty}R_T/T=0$, where $R_T$ is the cumulative regret
\[
R_T = \sum_{t=1}^T f(\xstar) - f(\x_t).
\]
This in turn implies a lower-bound on the convergence rate for the optimization problem.

Figures~\ref{fig:acquisition} and \ref{fig:acquisition2d} show how with the same GP posterior, different acquisition functions with different maxima are defined. Figure~\ref{fig:eivsucb} gives an example of how $\PI$, $\EI$ and $\GPUCB$ give rise to distinct sampling behaviour over time.

With several different parameterized acquisition functions in the literature, it is often unclear which one to use. Brochu \emph{et al.} \shortcite{Brochu:2010a} present one method of utility selection.   Instead of using a single acquisition function, they adopt a portfolio of acquisition functions governed by an online multi-armed bandit strategy, which almost always outperforms the best individual acquisition function of a suite of standard test problems.

\subsubsection{Maximizing the acquisition function}\label{sec:maxacquisition}

To find the point at which to sample, we still need to maximize the constrained objective $u(\x)$. \emph{Unlike the original unknown objective function, $u\func$ can be cheaply sampled}. We optimize the acquisition function using \emph{DIRECT} \cite{Jones:1993}, a deterministic, derivative-free optimizer. It uses the existing samples of the objective function to decide how to proceed to DIvide the feasible space into finer RECTangles. A particular advantage in active learning applications is that DIRECT can be implemented as an ``any-time'' algorithm, so that as long as the user is doing something else, it continues to optimize, and when interrupted, the program can use the best results found to that point in time.  Methods such as Monte Carlo and multistart have also been used, and seem to perform reasonably well \cite{Mockus:1994,Lizotte:2008}.

\subsection{Noise}\label{sec:noise}

The model we've used so far assumes that we have perfectly noise-free observations.  In real life, this is rarely possible, and instead of observing $f(\x)$, we can often only observe a noisy transformation of $f(\x)$.

The simplest transformation arises when $f(\x)$ is corrupted with Gaussian noise 
 $\epsilon \sim {\cal N}(0,\signoise^2)$ \cite{Rasmussen:2006}. If the noise is additive, we can easily add the noise distribution to the Gaussian distribution  ${\cal N}(0,\K)$ and define 
 \[
y_i = f(\x_i) + \epsilon_i.
\]
Since the mean is zero, this type of noise simply requires that we replace the kernel $\K$ with the following kernel for the noisy observations of $f(\cdot)$:
\begin{eqnarray}
    \K = \left[\begin{matrix}
            k(\x_1, \x_1) & \ldots & k(\x_1, \x_t)\\
            \vdots & \ddots & \vdots\\
            k(\x_t, \x_1) & \ldots & k(\x_t, \x_t)\\
        \end{matrix}\right] + \signoise^2 I\label{eqn:kernelmatrix}
\end{eqnarray}

This yields the predictive distribution:
\[
P(y_{t+1}|\data_{1:t},\x_{t+1}) = {\cal N} (\mu_t(\x_{t+1}), \sigma_t^2(\x_{t+1})+ \signoise^2),
\]
and the sufficient statistics
\begin{align*}
\mu_t(\x_{t+1})&=
\mathbf{k}^T [\mathbf{K} + \signoise^2 I ]^{-1} \y_{1:t} \\
\sigma_t^2(\x_{t+1})&= k(\x_{t+1},\x_{t+1}) - \mathbf{k}^T [\mathbf{K} + \signoise^2 I]^{-1}\mathbf{k}.
\nonumber
\end{align*}

In a noisy environment, we also change the definition of the incumbent in the PI and EI acquisition functions.  Instead of using the best observation, we use the distribution at the sample points, and define as the incumbent, the point with the highest expected value,
\[
\mu^+ = \argmax_{\x_i \in \x_{1:t}} \mu(\x_i).
\]  
This avoids the problem of attempting to maximize probability or expected improvement over an unreliable sample.  It is also possible to resample potential incumbents to get more reliable estimates of the values in a noisy environment \cite{Bartz-Beielstein:2005,Huang:2006,Hutter:2009}, a process sometimes called \emph{intensification}. Nonstationary noise models are also possible, such as autoregressive moving-average noise~\cite{Murray-Smith:2001} and heteroskedastic Gaussian noise~\cite{Goldberg:1998}.

\subsection{A brief history of Bayesian optimization}\label{sec:history}

The earliest work we are aware of resembling the modern Bayesian optimization approach is the early work of Kushner \shortcite{Kushner:1964}, who used Wiener processes for unconstrained one-dimensional problems.  Kushner's decision model was based on maximizing the probability of improvement (\Section \ref{sec:piei}).  He also included a parameter that controlled the trade-off between `more global' and `more local' optimization, in the same spirit as the exploration-exploitation trade-off.  A key difference is that in a (one-dimensional) Wiener process, the intervals between samples are independent, and Kushner was concerned with the problem of selecting from a finite set of intervals. Later work extended Kushner's technique to multidimensional optimization, using, for example, interpolation in a Delauney triangulation of the space \cite{Elder:1992} or projecting Wiener processes between sample points \cite{Stuckman:1988}.

Meanwhile, in the former Soviet Union, Mo\v{c}kus and colleagues developed a multidimensional Bayesian optimization method using linear combinations of Wiener fields.  This was first published in English as \cite{Mockus:1978}.  This paper also, significantly, describes an acquisition function that is based on myopic expected improvement of the posterior, which has been widely adopted in Bayesian optimization as the expected improvement function (\Section \ref{sec:piei}).  A more recent review of Mo\v{c}kus' approach is \cite{Mockus:1994}. 

At the same time, a large, related body of work emerged under the name \emph{kriging} (\Section \ref{sec:kriging}), in honour of the South African student who developed this technique at the University of the Witwatersrand \cite{Krige:1951}, though largely popularized by Matheron and colleagues (e.g. \cite{Matheron:1971}).  In kriging, the goal is interpolation of a random field via a linear predictor.  The errors on this model are typically assumed to \emph{not} be independent, and are modelled with a Gaussian process.

More recently, Bayesian optimization using Gaussian processes has been successfully applied to derivative-free optimization and experimental design, where it is called Efficient Global Optimization, or \emph{EGO} 
(\Section \ref{sec:ed}).

There exist several consistency proofs for this algorithm in the one-dimensional setting \cite{Locatelli:1997} and one for a simplification of the algorithm using simplicial partitioning in higher dimensions \cite{Zilinskas:2002}. The convergence of the algorithm using multivariate Gaussian processes has been recently established in \cite{Vasquez:2008}.

\subsection{Kriging}\label{sec:kriging}

Kriging has been used in geostatistics and environmental science since the 1950s and remains important today.  We will briefly summarize the connection to Bayesian optimization here.  More detailed examinations can be found in, for example, \cite{Stein:1999,Sasena:2002,Diggle:2007}.  This section is primarily drawn from these sources.

In many modelling techniques in statistics and machine learning, it is assumed that samples drawn from a process with independent, identically distributed residuals, typically, $\varepsilon \sim \mathcal{N}(0, \signoise^2)$:
\begin{eqnarray*}
y(\x) = f(\x) + \varepsilon
\end{eqnarray*}

In kriging, however, the usual assumption is that errors are \emph{not} independent, and are, in fact, spatially correlated: where errors are high, it is expected that nearby errors will also be high.  Kriging is a combination of a linear regression model and a stochastic model fitted to the residual errors of the linear model.  The residual is modelled with a zero-mean Gaussian process, so $\varepsilon$ is actually parameterized by $\x$: $\varepsilon(\x) \sim \mathcal{N}(0, \sigma^2(\x))$.

The actual regression model depends on the type of kriging.  In \emph{simple kriging}, $f$ is modelled with the zero function, making it a zero-mean GP model.  In \emph{ordinary kriging}, $f$ is modelled with a constant but unknown function.  \emph{Universal kriging} models $f$ with a polynomial of degree $k$ with bases $m$ and coefficients $\beta$, so that 
\[
y(\x) = \sum_{j=1}^k \beta_j m_j(\x) + \varepsilon(\x).
\]
Other, more exotic types of kriging are also used.

Clearly, kriging and Bayesian optimization are very closely related.  There are some key differences in practice, though.  In Bayesian optimization, models are usually fit through maximum likelihood.  In kriging, models are usually fit using a \emph{variogram}, a measure of the average dissimilarity between samples versus their separation distance.  Fitting is done using least squares or similar numerical methods, or interactively, by an expert visually inspecting the variogram plot with specially-designed software.  Kriging also often restricts the prediction model to use only a small number of neighbours, making it fit locally while ignoring global information.  Bayesian optimization normally uses all the data in order to learn a global model.

\subsection{Experimental design}\label{sec:ed}

Kriging has been applied to experimental design under the name \emph{DACE}, after ``Design and Analysis of Computer Experiments'', the title of a paper by Sacks \emph{et al.}\ \shortcite{Sacks:1989} (and more recently a book by Santner \emph{et al.}\ \shortcite{Santner:2003}).  In DACE, the regression model is a best linear unbiased predictor (BLUP), and the residual model is a noise-free Gaussian process.  The goal is to find a design point or points that optimizes some criterion.  

The ``efficient global optimization'', or \emph{EGO}, algorithm is the combination of DACE model with the sequential expected improvement (\Section \ref{sec:piei}) acquisition criterion.  It was published in a paper by Jones \emph{et al.}\ \shortcite{Jones:1998} as a refinement of the \emph{SPACE} algorithm  (Stochastic Process Analysis of Computer Experiments) \cite{Schonlau:1997}.  Since EGO's publication, there has evolved a body of work devoted to extending the algorithm, particularly in adding constraints to the optimization problem \cite{Audet:2000,Sasena:2002,Boyle:2007}, and in modelling noisy functions \cite{Bartz-Beielstein:2005,Huang:2006,Hutter:2009,Hutter:2009a}.



In so-called ``classical'' experimental design, the problem to be addressed is often to learn the parameters $\zetav$ of a function $g_{\zetav}$ such that
\[
y_i = g_{\zetav} (\x_i) + \varepsilon_i, \forall i \in 1, \ldots, t
\]
with noise $\varepsilon_i$ (usually Gaussian) for scalar output $y_i$.  $\x_i$ is the $i^\mathrm{th}$ set of experimental conditions.  Usually, the assumption is that $g_{\zetav}$ is linear, so that
\[
y_i = \zetav^T \x_i + \varepsilon_i.
\]
An experiment is represented by a design matrix $\X$, whose rows are the inputs $\x_{1:t}$.  If we let $\varepsilon \sim \mathcal{N}(0, \sigma)$, then for the linear model, the variance of the parameter estimate $\widehat{\zetav}$ is
\[
    Var(\widehat{\zetav}) = \sigma^2 (\X^T \X)^{-1},
\]
and for an input $\x_i$, the prediction is
\[
    Var(\widehat{y}_t) = \sigma^2 \x_i^T (\X^T \X)^{-1} \x_i.
\]

An optimal design is a design matrix that minimizes some characteristic of the inverse moment matrix $(\X^T \X)^{-1}$.  Common criteria include A-optimality, which minimizes the trace; D-optimality, which minimizes the determinant; and E-optimality, which minimizes the maximum eigenvalue.


Experimental design is usually non-adaptive: the entire experiment is designed before data is collected.  However, \emph{sequential design} is an important and active subfield (e.g. \cite{Williams:2000,Busby:2009}.

\subsection{Active learning}

Active learning is another area related to Bayesian optimization, and of particular relevance to our task.  Active learning is closely related to experimental design and, indeed, the decision to describe a particular problem as active learning or experimental design is often arbitrary.  However, there are a few distinguishing characteristics that most (but by no means all) active learning approaches use.

\begin{itemize}
    \item Active learning is most often \emph{adaptive}: a model exists that incorporates all the data seen, and this is used to sequentially select candidates for labelling.  Once labelled, the data are incorporated into the model and new candidates selected.  This is, of course, the same strategy used by Bayesian optimization.
    \item Active learning employs an oracle for data labelling, and this oracle is very often a human being, as is the case with interactive Bayesian optimization.
    \item Active learning is usually concerned with selecting candidates from a finite set of available data (\emph{pool-based} sampling).  Experimental design and optimization are usually concerned with continuous domains.
    \item Finally, active learning is usually used to learn a model for classification, or, less commonly, regression.  Usually the candidate selection criterion is the maximization of some informativeness measure, or the minimization of uncertainty.  These criteria are often closely related to the alphabetic criteria of experimental design.
\end{itemize}

An excellent recent overview of active learning is~\cite{Settles:2010}.  We are particularly interested in active learning because it often uses a human oracle for label acquisition.  An example using GPs is the object categorization of Kapoor \emph{et al.}\ \shortcite{Kapoor:2007}.  In this work, candidates are selected from a pool of unlabelled images so as to maximize the margin of the GP classifier and minimize the uncertainty.  Osborne \emph{et al.} \shortcite{Osborne:2010a} also use active learning in a Gaussian process problem, deciding when to sample variables of interest in a sequential sampling problem with faults and changepoints.  Chu and Ghahramani \shortcite{Chu:2005b} briefly discuss how active learning could be used for \emph{ranking} GPs, by selecting sample points that maximize entropy gained with a new preference relation.

Interesting recent work with GPs that straddles the boundary between active learning and experimental design is the sensor placement problem of Krause \emph{et al.}\ \shortcite{Krause:2008}.  They examine several criteria, including maximum entropy, and argue for using mutual information.  Ideally, they would like to simultaneously select a set of points to place the entire set of sensors in a way that maximizes the mutual information.  This is essentially a classical experimental design problem with maximum mutual information as the design criterion.  However, maximizing mutual information over a set of samples is NP-complete, so they use an active learning approach.  By exploiting the submodularity of mutual information, they are able to show that sequentially selecting sensor locations that greedily maximize mutual information, they can bound the divergence of the active learning approach from the experimental design approach.  This work influenced the $\GPUCB$ acquisition function (\Section \ref{sec:ucb}).

Finally, as an aside, in active learning, a common acquisition strategy is selecting the point of maximum uncertainty.  This is called \emph{uncertainty sampling} \cite{Lewis:1994}.  GPs have in the useful property that the posterior variance (interpreted as uncertainty) is independent of the actual observations!  As a result, if this is the criterion, the entire active learning scheme can be designed before a single observation are made, making adaptive sampling unnecessary.

\subsection{Applications}\label{sec:priorapplications}

Bayesian optimization has recently begun to appear in the machine learning literature as a means of optimizing difficult black box optimizations.  A few recent examples include:

\begin{itemize}
    \item Lizotte \emph{et al.}\ \shortcite{Lizotte:2007,Lizotte:2008} used Bayesian optimization to learn a set of robot gait parameters that maximize velocity of a Sony AIBO ERS-7 robot.  As an acquisition function, the authors used maximum probability of improvement (\Section \ref{sec:piei}).  They show that the Bayesian optimization approach not only outperformed previous techniques, but used drastically fewer evaluations.
    \item Frean and Boyle \shortcite{Frean:2008} use Bayesian optimization to learn the weights of a neural network controller to balance two vertical poles with different weights and lengths on a moving cart.
    \item Cora's MSc thesis \shortcite{Cora:2008} uses Bayesian optimization to learn a hierarchical policy for a simulated driving task.  At the lowest level, using the vehicle controls as inputs and fitness to a course as the response, a policy is learned by which a simulated vehicle performs various activities in an environment.
    \item Martinez--Cantin \emph{et al.}\ \shortcite{Martinez--Cantin:2009} also applied Bayesian optimization to policy search.  In this problem, the goal was to find a policy for robot path planning that would minimize uncertainty about its location and heading, as well as minimizing the uncertainty about the environmental navigation landmarks.
    \item Hutter's PhD thesis \shortcite{Hutter:2009a} studies methods of automatically tuning algorithm parameters, and presents several sequential approaches using Bayesian optimization.
    \item The work of Osborne, Garnett \emph{et al.} \cite{Osborne:2010,Osborne:2010a,Garnett:2010a} uses Bayesian optimization to select the locations of a set of (possibly heterogenous) sensors in a dynamic system.  In this case, the samples are a function of the locations of the entire set of sensors in the network, and the objective is the root mean squared error of the predictions made by the sensor network.
\end{itemize}

\section{Bayesian Optimization for Preference Galleries}\label{sec:prefgalleries}

The model described above requires that each function evaluation have a scalar response.  However, this is not always the case. In applications requiring human judgement, for instance, preferences are often more accurate than ratings.  Prospect theory, for example, employs utility models based on relation to a reference point, based on evidence that the human perceptual apparatus is attuned to evaluate differences rather than absolute magnitudes \cite{Kahneman:1979,Tversky:1992}.  We present here a Bayesian optimization application based on discrete choice for a ``preference gallery'' application, originally presented in \cite{Brochu:2007b,Brochu:2007}.

In the case of a person rating the suitability of a procedurally-generated animation or image, each sample of valuation involves creating an instance with the given parameters and asking a human to provide feedback, which is interpreted as the function response.  This is a very expensive class of functions to evaluate!  Furthermore, it is in general impossible to even sample the function directly and get a consistent response from users.  Asking humans to rate an animation on a numerical scale has built-in problems---not only will scales vary from user to user, but human evaluation is subject to phenomena such as \emph{drift}, where the scale varies over time, \emph{anchoring}, in which early experiences dominate the scale \cite{Siegel:1988,Payne:1993}.  However, human beings \emph{do} excel at comparing options and expressing a preference for one over others \cite{Kingsley:2006}.  This insight allows us to approach the optimization function in another way.  By presenting two or more realizations to a user and requiring only that they indicate preference, we can get far more robust results with much less cognitive burden on the user \cite{Kendall:1975}.  While this means we can't get responses for a valuation function directly, we model the valuation as a latent function, inferred from the preferences, which permits a Bayesian optimization approach.

Probability models for learning from discrete choices have a long history in psychology and econometrics  \cite{Thurstone:1927,McFadden:1980,Stern:1990}. They have been studied extensively, for example, in rating chess players, and the Elo system \cite{Elo:1978} was adopted by the World Chess Federation FIDE to model the probability of one player beating another.  It has since been adopted to many other two-player games such as Go and Scrabble, and, more recently, online computer gaming \cite{Herbrich:2006}.

Parts of \Section \ref{sec:probit} are based on \cite{Chu:2005a}, which presents a preference learning method using probit models and Gaussian processes. They use a Thurstone--Mosteller model (below), but with an innovative nonparametric model of the valuation function.

\subsection{Probit model for binary observations}\label{sec:probit}

The \emph{probit model} allows us to deal with binary observations of $f(\cdot)$ in general. That is, every time we try a value of $\x$, we get back a binary variable, say either zero or one. From the binary observations, we have to infer the latent function $f(\cdot)$. In order to marry the presentation in this section to the user modeling applications discussed later, we will introduce probit models in the particular case of preference learning.

Assume we have shown the user $M$ pairs of items from a set of $N$ instances. In each case, the user has chosen which item she likes best. The data set therefore consists of the ranked pairs: 
\[
\data = \{\rv_i \succ \cv_i ; \; \; i = 1, \ldots, M\},
\]
where the symbol $\succ$ indicates that the user prefers $\rv$ to $\cv$.  We
 use $\x_{1:t}$ to denote the $t$ distinct elements in the training data. That is, $\rv_i$ and $\cv_i$ correspond to two elements of $\x_{1:t}$. $\rv_i \succ \cv_i$ can be interpreted as a binary variable that takes value 1 when $\rv_i$ is preferred to $\cv_i$ and is 0 otherwise.

In the probit approach, we model the value functions $v\func$ for items $\rv$ and $\cv$ as follows: 
\begin{eqnarray}
v(\rv_i) &=& f(\rv_i) + \varepsilon\nonumber\\
v(\cv_i) &=& f(\cv_i) + \varepsilon,\label{eq:utility}
\end{eqnarray}
where the noise terms are Gaussian: $\varepsilon \sim {\cal N}(0, \signoise^2)$. Following \cite{Chu:2005a}, we assign a nonparametric Gaussian process prior to the unknown mean valuation: $f(\cdot) \sim \mathcal{GP}(0,K(\cdot,\cdot))$. That is, at the $t$ training points: 
\[
P(\f) =|2\pi\K|^{-\frac{1}{2}}\exp\left(-\frac{1}{2}\f^T\K^{-1}\f\right),
\]
where $\f = \{f(\x_1), f(\x_2), \ldots, f(\x_t)\}$.

Random utility models such as~(\ref{eq:utility}) have a long and influential history in psychology and the study of individual choice behaviour in economic markets. Daniel McFadden's Nobel Prize speech \cite{McFadden:2001} provides a glimpse of this history. Many more comprehensive treatments appear in classical economics books on discrete choice theory. 

Under our Gaussian utility models, the probability that item $\rv$ is preferred to item $\cv$ is given by:
\begin{eqnarray*}
P(\rv_i \succ \cv_i |f(\rv_i),f(\cv_i)) &=& P(v(\rv_i) > v(\cv_i)|f(\rv_i),f(\cv_i)) \\
&=& P(\varepsilon - \varepsilon < f(\rv_i) - f(\cv_i)) \\
&=& \Phi(Z_i),
\end{eqnarray*}
where
\[
Z_i =\frac{f(\rv_i) - f(\cv_i)}{\sqrt{2}\signoise}
\]
and $\Phi\func$ is the CDF of the standard normal distribution.  This model, relating binary observations to a continuous latent function, is known as the Thurstone-Mosteller law of comparative judgement \cite{Thurstone:1927,Mosteller:1951}. In statistics it goes by the name of binomial-probit regression. Note that one could also easily adopt a logistic (sigmoidal) link function $\varphi\left( Z_i\right) =\left( 1+\exp\left( -Z_i\right) \right) ^{-1}$. In fact, such choice is known as the Bradley-Terry model \cite{Stern:1990}. If the user had more than two choices one could adopt a polychotomous regression model \cite{holmes:2006}. This multi-category extension would, for example, enable the user to state no preference, or a degree of preference for any of the two items being presented. 

Note that this approach is related to, but distinct from the binomial logistic-linear model used in geostatistics \cite{Diggle:1998}, in which the responses $y$ represent the outcomes of Bernoulli trials which are conditionally independent given the model (i.e., the responses are binary observations $y_t \in \{0,1\}$ for $\x_t$, rather than preference observations between $\{\rv_t, \cv_t\}$).

Our goal is to estimate the posterior distribution of the latent utility function given the discrete data. That is, we want to maximize
\[
   P(\f|\data) \propto P(\f) \prod_{i=1}^{M} P(\rv_i \succ \cv_i | f(\rv_i), f(\cv_i)).
\]

Although there exist sophisticated variational and Monte Carlo methods for approximating this distribution, we favour a simple strategy: Laplace approximation.  The Laplace approximation follows from Taylor-expanding the log-posterior about a set point $\widehat{\f}$:
\[
   \log P(\f|\data) = \log P(\widehat{\f}|\data) + \g^T (\f-\widehat{\f}) - \frac{1}{2} (\f-\widehat{\f})^T \Hm (\f-\widehat{\f}), 
\]
where $\g = \nabla_{\f}\log P(\f|\data)$ and $\Hm = - \nabla_{\f}\nabla_{\f}\log P(\f|\data)$. At the mode of the posterior ($\widehat{\f} = \fmap$), the gradient $\g$ vanishes, and we obtain:
\[
   P(\f|\data) \approx P(\widehat{\f}|\data) \exp \left[ - \frac{1}{2} (\f-\widehat{\f})\Hm (\f-\widehat{\f})\right] 
\]
In order to obtain this approximation, we need to compute the maximum a posteriori (MAP) estimate $\fmap$, the gradient $\g$ and the information matrix $\Hm$.

The gradient is given by:
\begin{eqnarray}
   \g &=& \nabla_{\f} \log P(\f|\data) \nonumber \\
   &=& \nabla_{\f} \left[const -\frac{1}{2}\f^T\K^{-1}\f + \sum_{i=1}^{M} \log \Phi(Z_i) \right] \nonumber \\
   &=& - \K^{-1}\f + \nabla_f\left[ \sum_{i=1}^{M} \log \Phi(Z_i)\right] = - \K^{-1}\f + \bv, \nonumber
\end{eqnarray}
where the $j$-th entry of the $N$-dimensional vector $\bv$ is
given by:
\[
   b_j = \frac{1}{\sqrt{2}\signoise} \sum_{i=1}^{M} \frac{\phi(Z_i)}{\Phi(Z_i)} \left[  \frac{\partial}{\partial f(\x_j)}   (f(\rv_i) - f(\cv_i))  \right],
\]
where $\phi\func$ denotes the PDF of the standard normal distribution. Clearly, the derivative $h_i(\x_j) = \frac{\partial}{\partial f(\x_j)} (f(\rv_i) - f(\cv_i))$ is 1 when $\x_j=\rv_i$, -1 when $\x_j=\cv_i$ and 0 otherwise. Proceeding to compute the second derivative, one obtains the Hessian: $\Hm = \K^{-1} + \C$, where the matrix $\C$ has entries
\begin{eqnarray}
   \C_{m,n} &=& - \frac{\partial^2}{\partial f(\x_m)\partial f(\x_n)}\sum_{i=1}^{M} \log \Phi(Z_i) \nonumber \\ &=&\frac{1}{{2}\sigma^2}\sum_{i=1}^{M}h_i(\x_m)h_i(\x_n)\left[\frac{\phi(Z_i)}{\Phi^2(Z_i)}+\frac{\phi^2(Z_i)}{\Phi(Z_i)}Z_i\right]\nonumber
\end{eqnarray}

\begin{figure}[t!]
\begin{minipage}{3cm}
    \begin{tabular}{c}
        \hline
        \textbf{preferences}\\
        \hline
        $0.2 \succ 0.1$\\
        $0.35 \succ 0.5$\\
        $0.2 \succ 0.35$\\
        $0.2 \succ 0.6$\\
        $0.8 \succ 0.7$\\
        \hline
    \end{tabular}
\end{minipage}
\begin{minipage}{9cm}
    \includegraphics[width=\textwidth]{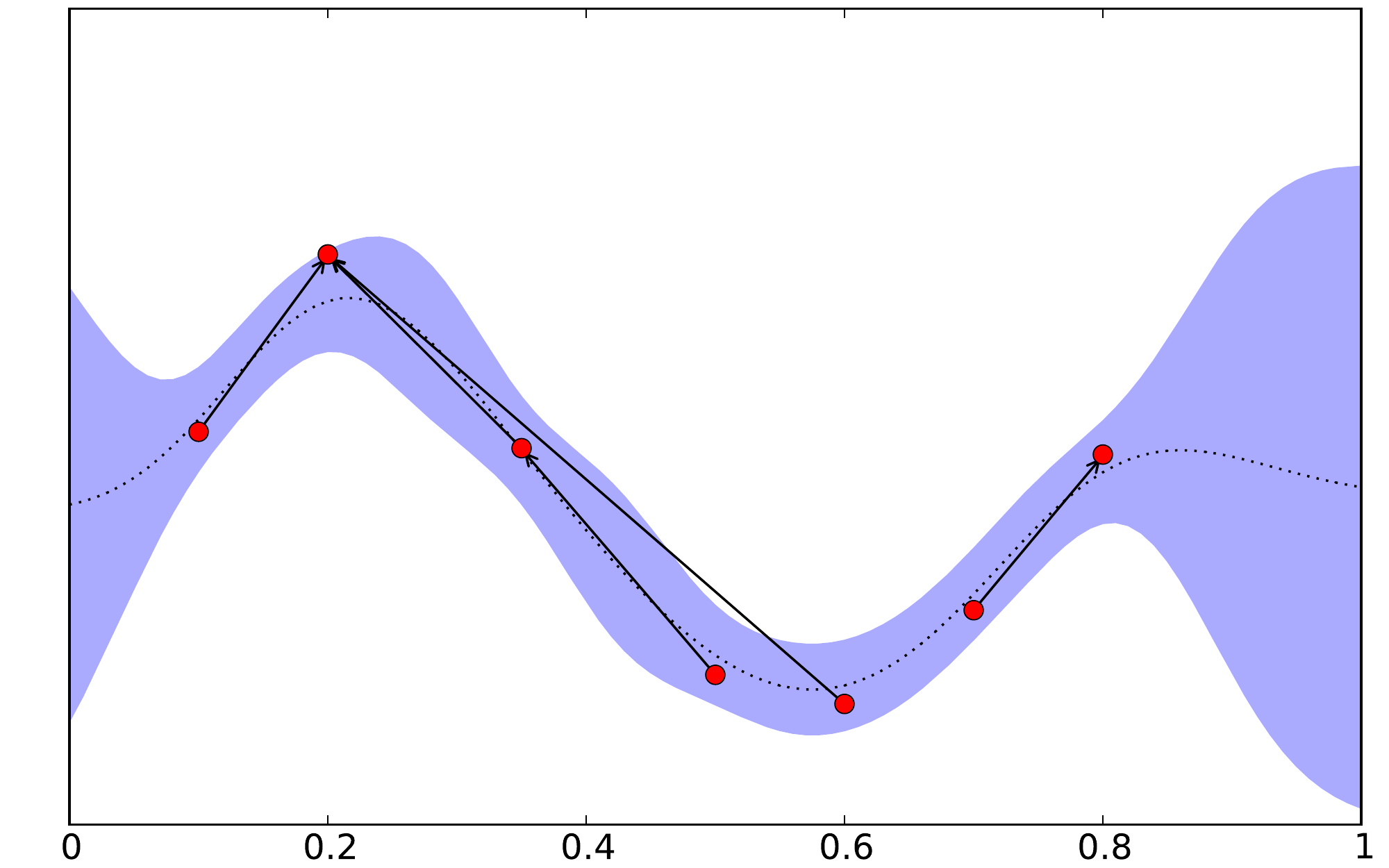}
\end{minipage}
 \caption[Example of a set of preference relations used to infer a GP on a toy problem.]{\capstyle{Example of a set of preference relations (left table) used to infer a GP (right plot) on a toy problem.  The preferences are indicated as a set of preferences between two points in the space, which serves as input to a function that finds a Gaussian process that takes into account all the available preference information, as well as prior information on the smoothness and noise.}}
 \label{fig:pref1d}
\end{figure}

The Hessian is a positive semi-definite matrix.  Hence, one can find the MAP estimate with a simple Newton--Raphson recursion:
\[
   \f^{\operatorname{new}} = \f^{\operatorname{old}} - \Hm^{-1}\g \left.\right|_{\f=\f^{\operatorname{old}}}.
\]
At $\f = \fmap$, we have
\[
   P(\f|\data) \approx {\cal N} \left(\K\bv,(\K^{-1}+\C)^{-1}\right).
\]
with $\bv = \K^{-1} \fmap$.  The goal of our derivation, namely the predictive distribution $P(f_{t+1}|\data)$, follows by straightforward convolution of two Gaussians:
\begin{eqnarray}
P(f_{t+1}|\data) &=& \int P(f_{t+1}|\fmap) P(\fmap|\data) d\fmap \nonumber \\
&\propto& {\cal N} (\kv^{T}\K^{-1}\fmap,k(\x_{t+1},\x_{t+1})-\kv^{T}(\K +
 \C^{-1})^{-1}\kv). \nonumber
\end{eqnarray}
An example of the procedure on a toy 1D problem is shown in Figure~\ref{fig:pref1d}.  We can see that the inferred model explains the observed preferences, while also adhering to prior information about smoothness and noise.

\subsection{Application: Interactive Bayesian optimization for material design}\label{sec:brdf}

\begin{figure*}[h!]
\centering
$
\begin{array}{ll}
    \includegraphics[height=3cm]{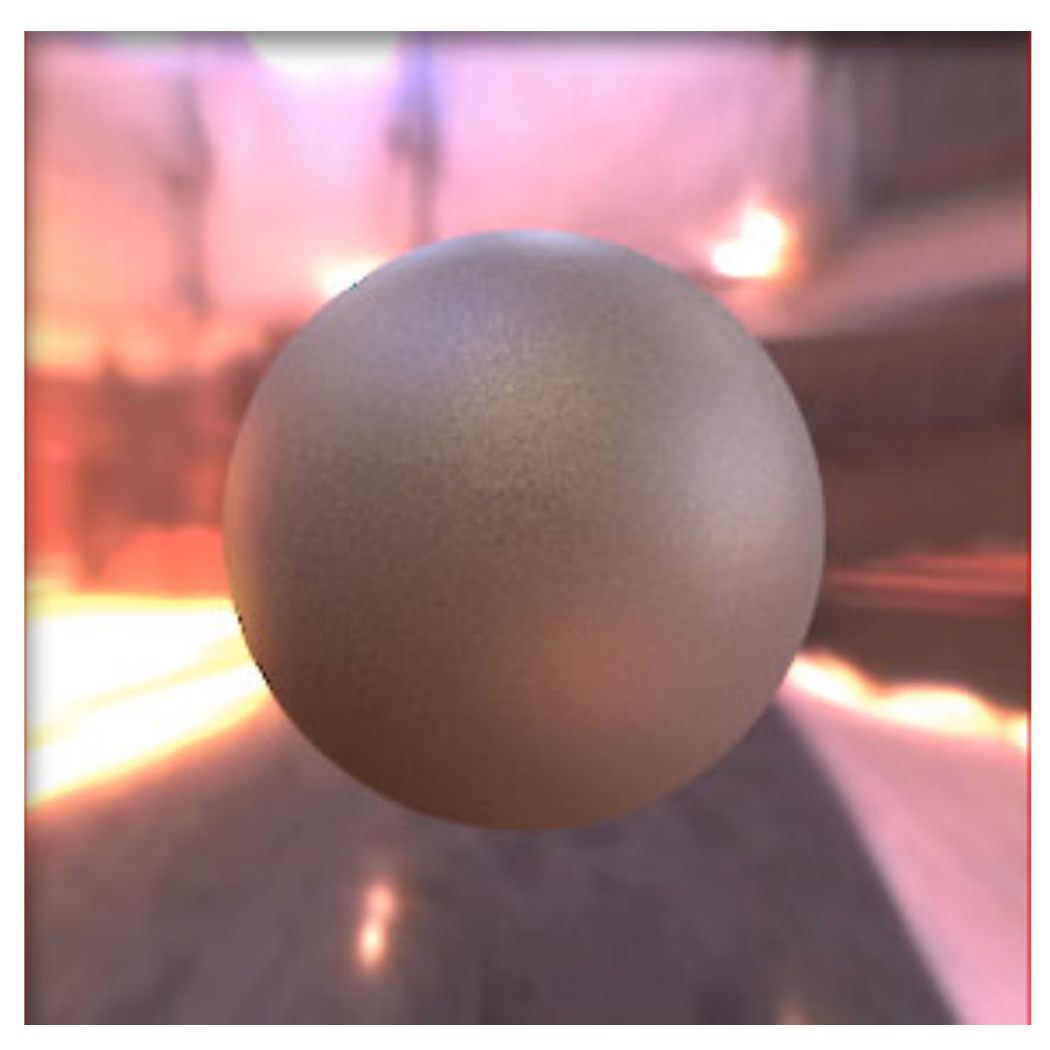}~&~
    \includegraphics[height=3cm]{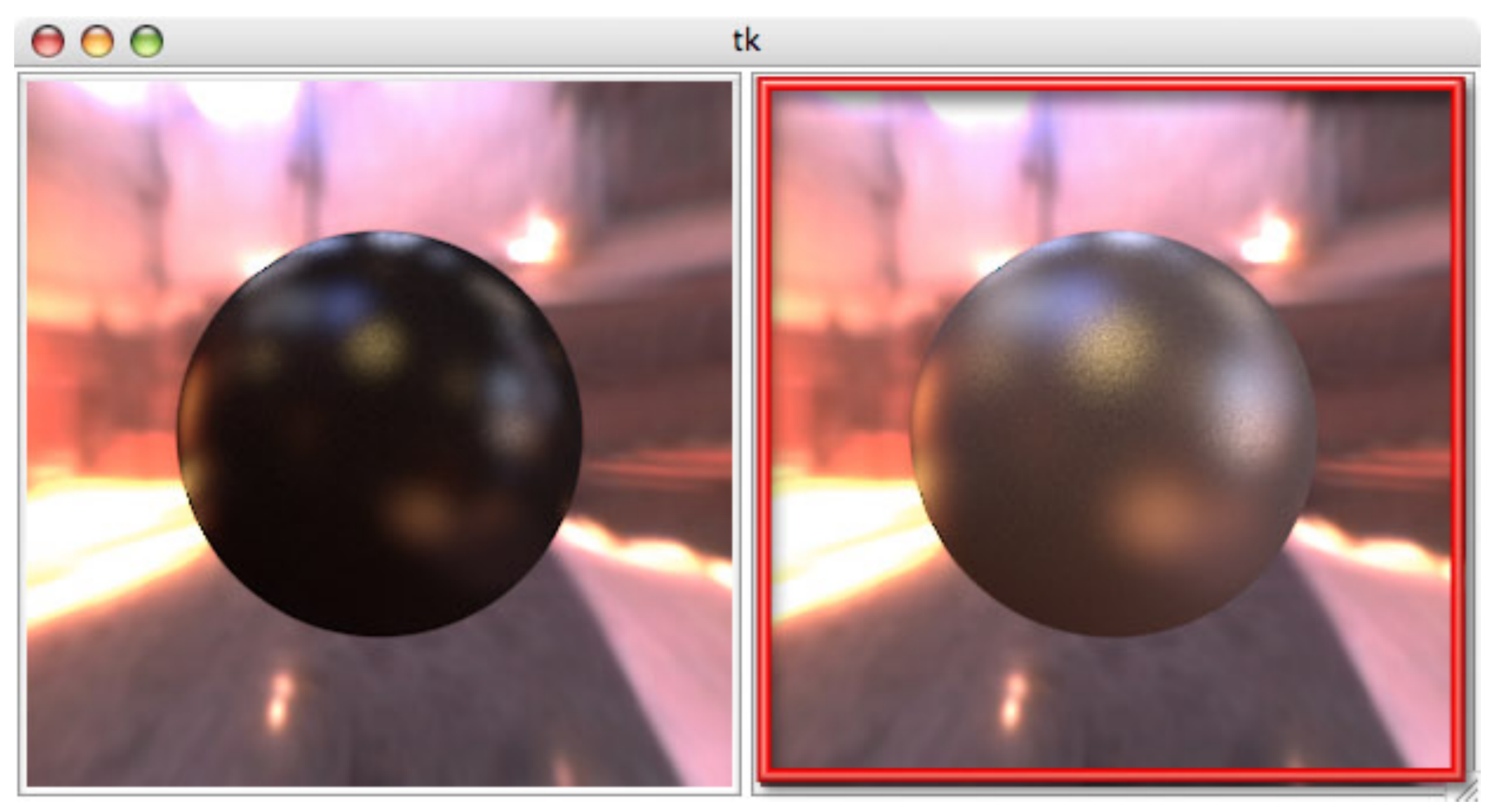}\\
    Target ~&~ 1.\\
    \vspace{-.2cm}\\
    ~&~\includegraphics[height=3cm]{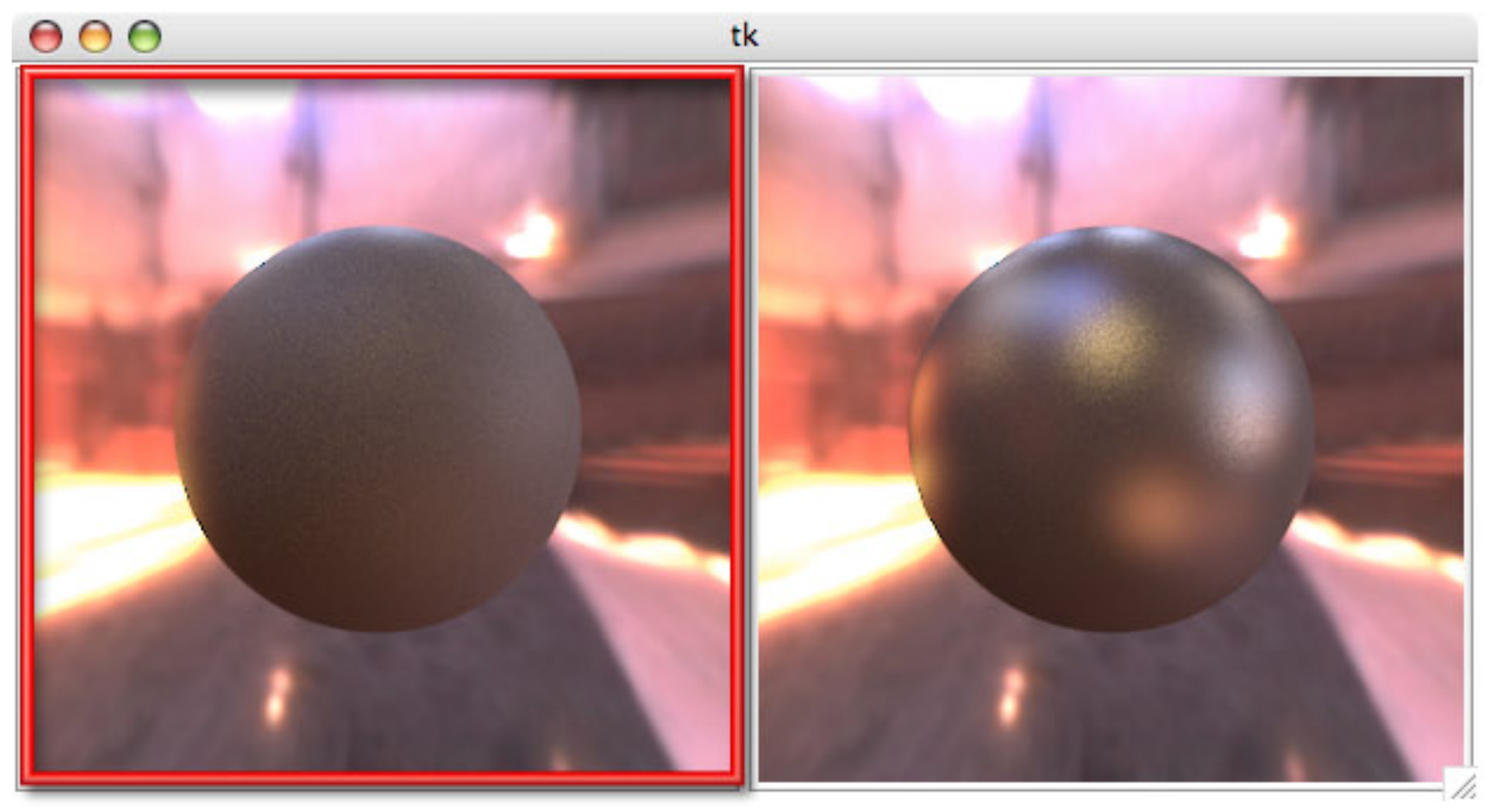}\\
     ~&~ 2.\\
     \vspace{-.2cm}\\
    ~&~\includegraphics[height=3cm]{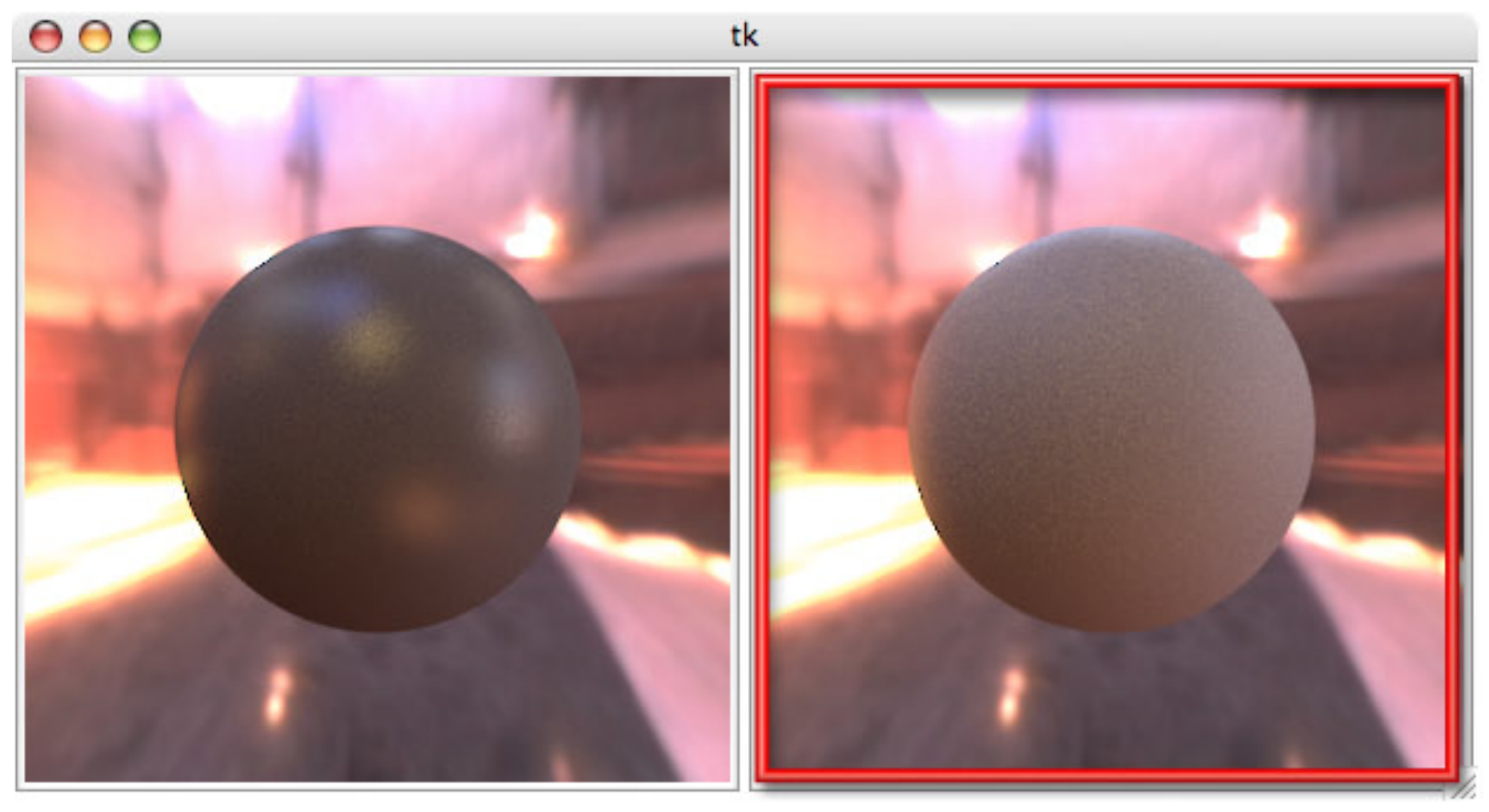}\\
    ~&~ 3.\\
    \vspace{-.3cm}\\
    ~&~\includegraphics[height=3cm]{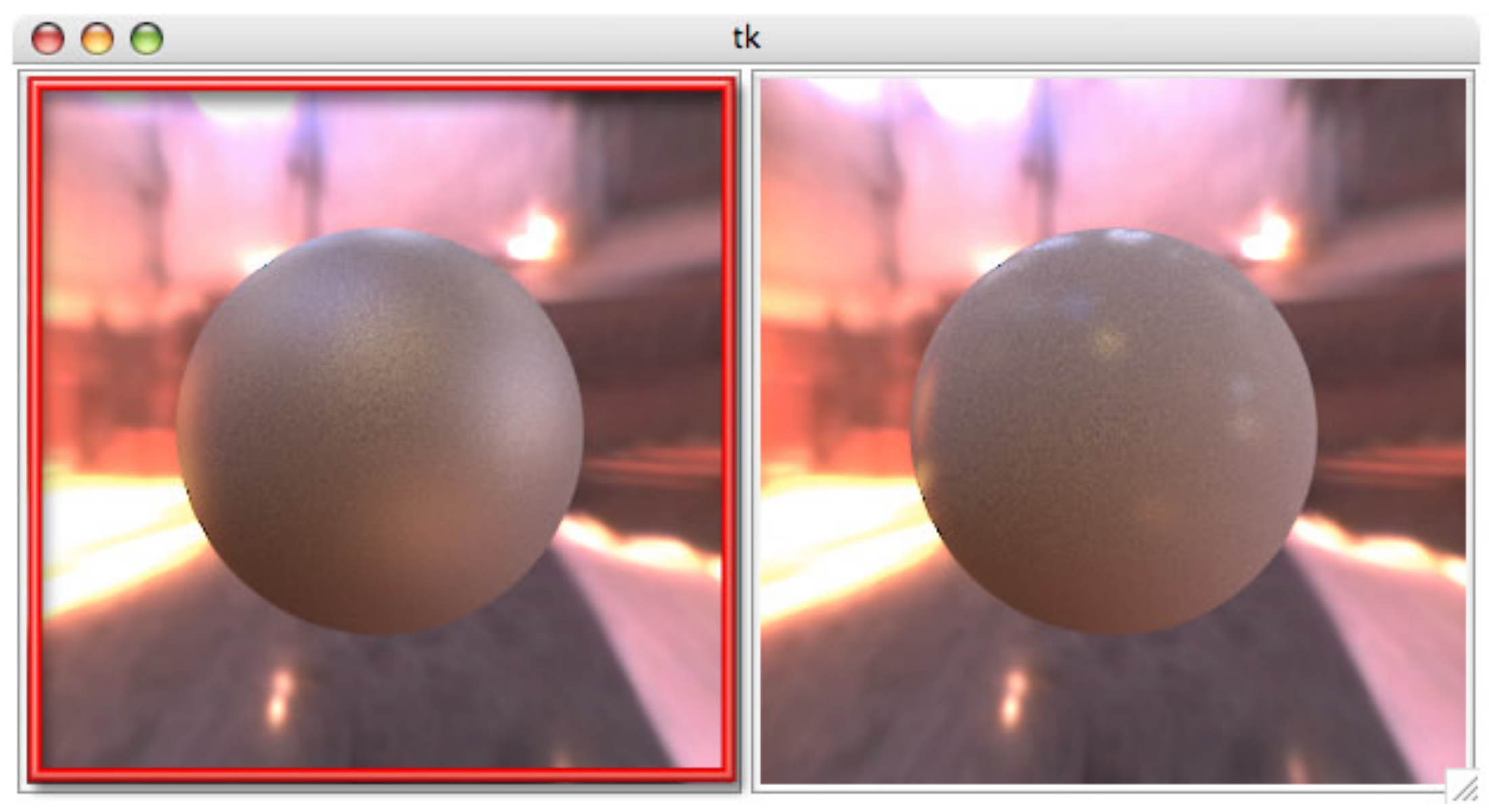}\\
     ~&~ 4.
\end{array} 
$
  \caption[A shorter-than-average but otherwise typical run of the
  BRDF preference gallery tool.]{\capstyle{A shorter-than-average but otherwise typical run of the
BRDF preference gallery tool.  At each (numbered) iteration, the user is
provided with two images generated with parameter instances and
indicates the one they think most resembles the target image (top-left) they are looking for.  The boxed images are the user's selections at each iteration.}}
  \label{fig:brdfrun}
\end{figure*}

Properly modeling the appearance of a material is a necessary component of
realistic image synthesis. The appearance of a material is formalized
by the notion of the Bidirectional Reflectance Distribution Function
(BRDF). In computer graphics, BRDFs are most often specified using
various analytical models. Analytical models that are of interest to
realistic image synthesis are the ones that observe the physical laws
of reciprocity and energy conservation while typically also exhibiting
shadowing, masking and Fresnel reflectance phenomenon. Realistic
models are therefore fairly complex with many parameters that need to be
adjusted by the designer for the proper material
appearance. Unfortunately these parameters can interact in
non-intuitive ways, and small adjustments to certain settings may
result in non-uniform changes in the appearance. This can make the
material design process quite difficult for the artist end user, who is not expected to be an expert in the field, but who knows the look that she desires for a
particular application without necessarily being interested in
understanding the various subtleties of reflection. We attempt to deal with this using a preference gallery approach, in which users are simply required to view two or more images rendered with different material properties and indicate which they prefer, in an iterative process.


We use the interactive Bayesian optimization model with probit responses on an example gallery
application for helping users find a BRDF. For the purposes of this
example, we limit ourselves to isotropic materials and ignore
wavelength dependent effects in reflection. Our gallery demonstration presents the
user with two BRDF images at a time. We start with four predetermined
queries to ``seed'' the parameter space, and after that use the
learned model to select gallery images. The GP model is updated after
each preference is indicated. We use parameters of real measured
materials from the MERL database for seeding the parameter space, but can draw arbitrary parameters after that. 

By querying the user with a paired comparison, one can estimate statistics of the valuation function at the query point, but only at considerable expense. Thus, we wish to make sure that the samples we do draw will generate the maximum possible improvement.

Our method for achieving this goal iterates over the following steps: {\small
\begin{enumerate}
   \item \textbf{Present the user with a new set of instances and record preferences from the user}: Augment the training set of paired choices with the new user data.
   \item \textbf{Infer the valuation function}: Here we use a Thurstone--Mosteller model with Gaussian processes. See \Section \ref{sec:probit} for details.  Note that in this application, the valuation function is the objective of Bayesian optimization.  We will use the terms interchangeably.
   \item \textbf{Optimize the acquisition function of the valuation to obtain the query points for the next gallery}: Methods for selecting a set of instances are described in \Section \ref{sec:user_study1}.
\end{enumerate}}

\subsubsection{User Study}\label{sec:user_study1}

To evaluate the performance of our application, we have run a simple
user study in which the generated images are restricted to a subset of
38 materials from the MERL database that we deemed to be representative of the appearance space of the measured materials.  The user is given the task of finding a single randomly-selected image from that set by indicating preferences.  Figure~\ref{fig:brdfrun} shows a typical user run, where we ask the user to use the preference gallery to find a provided target image.  At each step, the user need only indicate the image they think looks most like the target.  This would, of course, be an unrealistic scenario if we were to be evaluating the application from an HCI stance, but here we limit our attention to the model, where we are interested here in demonstrating that with human users maximizing valuation is preferable to learning the entire latent function.  

Using five subjects, we ran 50 trials each of three different methods of selecting sample pairs\footnote{An empirical study of various methods on a variety of test functions, and a discussion of why these were selected can be found in \cite{Brochu:2007b}.}.  In all cases, one of the pair is the incumbent $\argmax_{\x_i} \mu(\x_i), \x_i \in \x_{1:t}$.  The second is selected via one of the following methods:

\begin{itemize}
    \item \textbf{random}  The second point is sampled uniformly from the parameter domain $\mathcal{A}$.
    \item \textbf{$\mathbf{argmax_{\sigma}}$}  The second point is the point of highest uncertainty, $\argmax_\x \sigma(\x)$.
    \item \textbf{$\mathbf{argmax_{\EI}}$}  The second point is the point of maximum expected improvement, $\argmax_\x \EI(\x)$.
\end{itemize}
The results are shown in Table~\ref{tab:studyresults}.  $n$ is the number clicks required of the user to find the target image.  Clearly $\argmax_{\EI}$ dominates, with a mean $n$ less than half that of the competing algorithms.  Interestingly, selecting images using maximum variance does not perform much better than random.  We suspect that this is because $\argmax_{\sigma}$ has a tendency to select images from the corners of the parameter space, which adds limited information to the other images, whereas Latin hypercubes at least guarantees that the selected images fill the space.

\begin{table}[t!] 
\centering
\begin{tabular}{|l|cc|}
	\hline
	\textbf{algorithm} & \textbf{trials} &\textbf{ $n$ (mean $\pm$ std)}\\
	\hline
	\textbf{random} & 50 & 18.40 $\pm$ 7.87\\
	$\mathbf{argmax_{\sigma}}$ & 50 & 17.87 $\pm$ 8.60\\
	$\mathbf{argmax_{\EI}}$ & 50 & 8.56 $\pm$ 5.23\\
	\hline
\end{tabular}
\caption[Results of the user study on the BRDF gallery.]{\capstyle{Results of the user study on the BRDF gallery.}}\label{tab:studyresults}
\end{table}

\section{Bayesian Optimization for Hierarchical Control}\label{sec:hiercontrol}

In general, problem solving and planning becomes easier when it is broken down into subparts. Variants of functional hierarchies appear consistently in video game AI solutions, from behaviour trees, to hierarchically decomposed agents (teams vs. players), implemented by a multitude of customized hierarchical state machines. The benefits are due to isolating complex decision logic to fairly independent functional units (tasks). The standard game AI development process consists of the programmer implementing a large number of behaviours (in as many ways as there are published video games), and hooking them up to a more manageable number of tuneable parameters. We present a class of algorithms that attempt to bridge the gap between game development, and general reinforcement learning. They reduce the amount of hand-tuning traditionally encountered during game development, while still maintaining the full flexibility of manually hard-coding a policy when necessary.

The Hierarchical Reinforcement Learning~\cite{Barto:2003} field models repeated decision making by structuring the policy into tasks (actions) composed of subtasks that extend through time (temporal abstraction) and are specific to a subset of the total world state space (state abstraction). Many algorithms have recently been developed, and are described further in Section~\ref{sec:HRL}.  The use of Bayesian optimization for control has previously been proposed by Murray-Smith and Sbarbaro \shortcite{Murray--Smith:2002}, and (apparently independently) by Frean and Boyle \shortcite{Frean:2008}, who used it for a control problem of balancing two poles on a cart.  This work did not involve a nonhierarchical setting, however.

The exploration policies typically employed in HRL research tend to be slow in practice, even after the benefits of state abstraction and reward shaping.  We demonstrate an integration of the MAXQ hierarchical task learner with Bayesian active exploration that significantly speeds up the learning process, applied to hybrid discrete and continuous state and action spaces. Section~\ref{sec:domain} describes an extended Taxi domain, running under The Open Racing Car Simulator~\cite{Wymann:2009}, a 3D game engine that implements complex vehicle dynamics complete with manual and automatic transmission, engine, clutch, tire, suspension and aerodynamic models.

\subsection{Hierarchical Reinforcement Learning}
\label{sec:HRL}

Manually coding hierarchical policies is the mainstay of video game AI development. The requirements for automated HRL to be a viable solution are it must be easy to customize task-specific implementations, state abstractions, reward models, termination criteria and it must support continuous state and action spaces. Out of the solutions investigated, MAXQ~\cite{Dietterich:2000} met all our requirements, and was the easiest to understand and get positive results quickly. The other solutions investigated include HAR and RAR~\cite{Ghavamzadeh:2005} which extend MAXQ to the case of average rewards (rather than discounted rewards). The implementation of RAR is mostly the same as MAXQ, and in our experiments gave the same results. Hierarchies of Abstract Machines (HAM)~\cite{Parr:1998} and ALisp~\cite{Andre:2003} are an exciting new development that has been recently applied to a Real-Time-Strategy (RTS) game~\cite{Marthi:2005}. ALisp introduces programmable reinforcement learning policies that allows the programmer to specify choice points for the algorithm to optimize. Although the formulation is very nice and would match game AI development processes, the underlying solver based on HAMs flattens the task hierarchy by including the program's memory and call-stack into a new joint-state space, and solves this new MDP instead. It is less clear how to extend and implement per-task customized learning with this formulation. Even if this difficulty is surmounted, as evidenced by the last line in the concluding remarks of~\cite{Marthi:2005}, there is an imperative need for designing faster algorithms in HRL. This paper aims to address this need.

In our solution, we still require a programmer or designer to specify the task hierarchy. In most cases breaking a plan into sub-plans is much easier than coding the decision logic. With the policy space constrained by the task hierarchy, termination and state abstraction functions, the rate of learning is greatly improved, and the amount of memory required to store the solution reduces. The benefits of HRL are very dependant however on the quality of these specifications, and requires the higher-level reasoning of a programmer or designer. An automatic solution to this problem would be an agent that can learn how to program, and anything less than that will have limited applicability.

We can use Bayesian optimization to learn the relevant aspects of value functions by focusing on the most relevant parts of the parameter space.  In the work on this section, we use refer to the objective as the \emph{value} function, to be consistent with the HRL literature.

\subsubsection{Semi-MDPs}
Each task in an HRL hierarchy is a semi-Markov Decision Process~\cite{Sutton:1999}, that models repeated decision making in a stochastic environment, where the actions can take more than one time step. Formally, an SMDP is defined as a tuple: $\{S, A, P(s',N|s,a), R(s,a)\}$ where $S$ is the set of state variables, $A$ is a set of actions, $P(s',N|s,a)$ is the transition probability of arriving to state $s'$ in $N$ time steps after taking action $a$ in $s$, and $R(s,a)$ is the reward received. The solution of this process is a policy $\pi^*(s) \in A$, that selects the action with the highest expected discounted reward in each state. The function $V^*(s)$ is the value of state $s$ when following the optimal policy. Equivalently, the $Q^*(s,a)$ function stores the value of taking action $a$ in state $s$ and following the optimal policy thereafter. These quantities follow the classical Bellman recursions:

\begin{equation} 
   V^*(s) = \max_{a\in A} \left[ R(s,a) + \gamma \sum_{s',N} P(s', N|s,a) \gamma^N V^*(s')  \right]\nonumber
\end{equation}
\begin{equation}
   Q^*(s,a) =  R(s,a) + \gamma \sum_{s',N} P(s',N|s,a) \gamma^N V^*(s') 
\end{equation}

\subsubsection{Hierarchical Value Function Decomposition}

A task $i$ in MAXQ~\cite{Dietterich:2000} is defined as a tuple: $\{A_i, T_i(s), Z_i(s), \pi_i(s)\}$ where $s$ is the current world state, $A_i$ is a set of subtasks, $T_i(s) \in \{true,false\}$ is a termination predicate, $Z_i(s)$ is a state abstraction function that returns a subset of the state relevant to the current subtask, and $\pi_i(s) \in A_i$ is the policy learned by the agent (or used to explore during learning). Each task is effectively a separate, decomposed SMDP that has allowed us to integrate active learning for discrete map navigation with continuous low-level vehicle control. This is accomplished by decomposing the $Q$ function into two parts:
\begin{eqnarray}
   a &=& \pi_i(s)  \\
   Q^{\pi}(i,s,a) &=& V^\pi(a, s) + C^\pi(i, s, a) \nonumber \\
   C^{\pi}(i, s, a) &=& \displaystyle\sum_{s',N}P_i^\pi(s',N|s,a)\gamma^N Q^\pi(i,s',\pi_i(s')) \nonumber \\
   V^\pi(i,s) &=& \left\{
       \begin{array}{l l}
               Q^\pi(i,s,\pi_i(s)) & \mbox{if composite}\\
               \sum_{s'} P(s'|s,i)R(s'|s,i) & \mbox{if primitive}\\
       \end{array}
   \right. \nonumber
\end{eqnarray}

Here, $\gamma$ is the discount factor, $i$ is the current task, and $a$ is a child action given that we are following policy $\pi_i$. The $Q$ function is decomposed into two parts: the value of $V^\pi$ being the expected one step reward, plus $C^\pi$ which is the expected completion reward for $i$ after $a$ completes. $V$ is defined recursively, as the expected value of its child actions, or the expected reward itself if $i$ is a primitive (atomic) action. The MAXQ learning routine is a simple modification of the typical Q-learning algorithm. In task $i$, we execute subtask $a$, observe the new state $s'$ and reward $r$. If $a$ is primitive, we update $V(s,a)$, otherwise we update $C(i,s,a)$, with learning rate $\alpha \in (0,1)$:

\begin{eqnarray}
V(a,s) &=& (1-\alpha) \times V(a,s) + \alpha \times r \\
C(i,s,a) &=& (1-\alpha) \times C(i,s,a) + \alpha \times \max_{a'} Q(i,s',a') \nonumber \nonumber
\end{eqnarray}

An important consideration in HRL is whether the policy calculated is hierarchically or recursively optimal. Recursive optimality, satisfied by MAXQ and RAR, means that each subtask is locally optimal, given the optimal policies of the descendants. This may result in a suboptimal overall policy because the effects of tasks executed outside of the current task's scope are ignored. For example if there are two exits from a room, a recursively optimal policy would pick the closest exit, regardless of the final destination. A hierarchically optimal policy (computed by the HAR~\cite{Ghavamzadeh:2005} and HAM~\cite{Andre:2003} three-part value decompositions) would pick the exit to minimize total travelling time, given the destination. A recursively optimal learning algorithm however generalizes subtasks easier since they only depend on the local state, ignoring what would happen after the current task finishes. So both types of optimality are of value in different degrees for different cases. The MAXQ formulation gives a programmer or designer the ability to selectively enable hierarchical optimality by including the relevant state features as parameters to a task. However, it may be difficult to identify the relevant features, as they would be highly application specific.

\begin{figure}
    \centering
    \includegraphics[width=9cm]{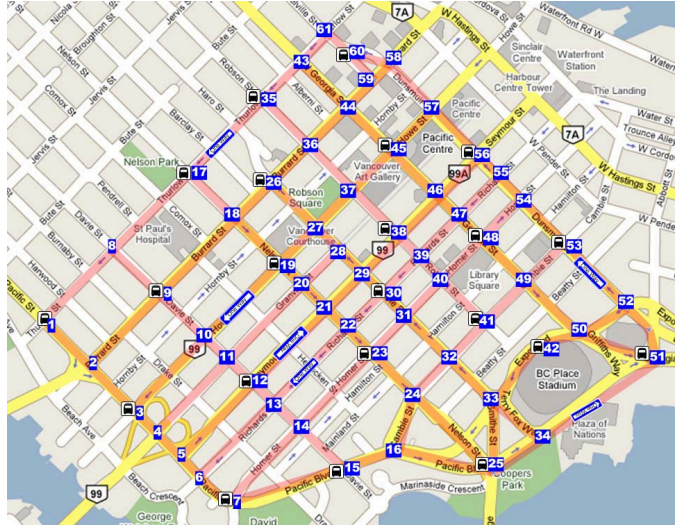}
    \caption{\capstyle{\textbf{City Experiment} uses a simplified map (orange overlay) roughly based on downtown Vancouver, and used by the TORCS simulator. Each waypoint is labeled, and pickup and dropoff locations are marked by the Taxi icons. One way streets are accounted for in the waypoint adjacency matrix. Source image care of Google Maps. }}\label{fig:vancity}
\end{figure}

\subsection{Application: The Vancouver Taxi Domain}
\label{sec:domain}

Our domain is a city map roughly based on a portion of downtown Vancouver, British Columbia, illustrated in Figure~\ref{fig:vancity}. The data structure is a topological map (a set of intersection nodes and adjacency matrix) with 61 nodes and 22 possible passenger pickup and drop-off locations. The total navigable area of the map is roughly 28 kilometers.

The state model includes both discrete variables used in the top layers of the task hierarchy, as well as continuous variables used by the \emph{Follow} task that tracks a trajectory, and are described in Table~\ref{table:states}. The original taxi domain~\cite{Dietterich:2000} is a 5x5 grid, with 4 possible pickup and dropoff destinations, and 6 actions (pickup, dropoff, and navigating North, South, East, West).

Table~\ref{table:policysize} makes a rough comparison between the size of our extended application and the original taxi domain. Ignoring the continuous trajectory states (including the \emph{Stopped} flag) and assuming the taxi hops from one intersection to an adjacent one in a single time step results in a fully discrete problem. A flat learning solution scales poorly, not only in terms of world samples required, but also in the size of the computed policy (if represented in a discrete table). The extended task hierarchy illustrated in Figure~\ref{fig:taskhierarchy} requires just a little bit more memory than the small 5x5 taxi domain.

\begin{table}[h]
   \caption{Comparing Domain Size}
\vspace{.5cm}
   \begin{tabular}{ll}
       \hline
       Domain & Size of final policy \\
       \hline

       5x5 Taxi Flat & $\sim 12,200~bytes$ \\
       Vancity Flat & $\sim 1,417,152~bytes$ \\
       Vancity Hierarchical & $\sim 18,668~bytes$ \\

       \hline
   \end{tabular}

\label{table:policysize}
\end{table}

\begin{table}[h]
\caption{States and Task Parameters}
\vspace{.5cm}

\label{table:states}
{

   \begin{tabular}{lll}
       \hline
       Name \TT \BB & Range/Units & Description \\
       \hline

       $TaxiLoc$ \TT &  \{0,1,..61\} & current taxi waypoint \#, or 0 if in \\
       &                             & transit between waypoints \\

       $PassLoc$ & \{0,1,..22\} & passenger waypoint \#, or 0 if in taxi \\

       $PassDest$ & \{1,2,..22\} & passenger destination waypoint \# \\

       $LegalLoad$ & \{$true,false$\} & true if taxi is empty and at  \\
       &                              & passenger, or loaded and at target\\

       $Stopped$  & \{$true,false$\} & indicates whether the taxi is at\\
       \BB &                               & a complete stop \\

       \hline

       $T$ \TT & \{1,2,..22\} & passenger location or destination\\
        &                      & parameter passed into \emph{Navigate} \\

       \emph{WP} \BB & \{1,2,..22\} & waypoint parameter adjacent to \\
       &                       & \emph{TaxiLoc} passed to \emph{Follow} \\

       \hline

       $Y_{err}$ \TT & meters & lateral error between desired point \\
       &                       & on the trajectory and vehicle \\

       $V_y$ & meters/second & lateral velocity (to detect drift) \\

       $V_{err}$ & meters/second & error between desired and real speed \\

       $\Omega_{err}$ & radians & error between trajectory angle and\\
       \BB & & vehicle yaw\\

       \hline
   \end{tabular}
}
\end{table}

%
%
\subsubsection{State Abstraction, Termination and Rewards}
\label{sec:hierarchy}

\begin{figure}
   \centering
   \includegraphics[width=10cm]{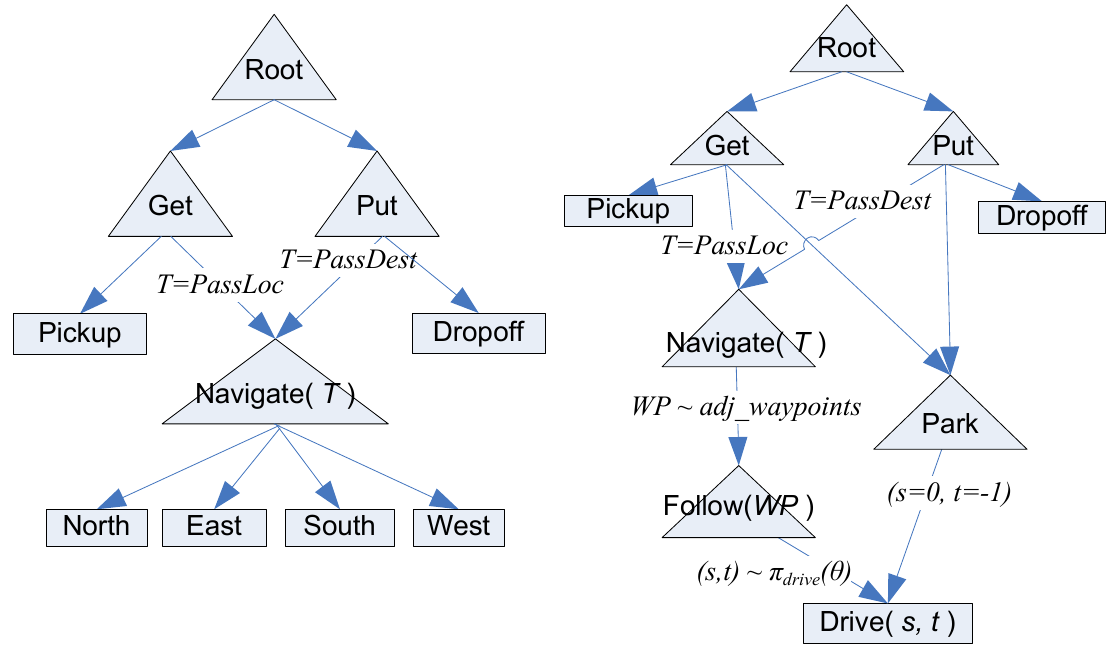}

   \caption{\capstyle{\textbf{Task Hierarchies.} Each composite task is a separate SMDP whose policy is optimal given the optimal policies of its subtasks (recursive optimality). Triangles are composite tasks, and rectangle are primitive actions. The hierarchy on the right simplifies learning by reusing policies for navigating form waypoint to waypoint, and the Navigation task only needs to learn the sequence of waypoints to get to the destination. For the continuous case, the discrete actions N/S/E/W are replaced by one continuous \emph{Drive(steer, throttle)} task, with driving parameters generated by the parameterized policy contained in the \emph{Follow} task.}}
   \label{fig:taskhierarchy}
\end{figure}

Figure~\ref{fig:taskhierarchy} compares the original task hierarchy, with our extended version that includes continuous trajectory following and a hard-coded \emph{Park} task. The state abstraction function filters out irrelevant states while computing the hash key for looking up and updating values of $V(s,a)$ and $C(i,s,a)$, where $s$ is the current state, $i$ is the current task, and $a$ is the child task. The \emph{Follow} task has been previously trained with the Active Policy optimizer from section~\ref{sec:apl} and the policy parameters fixed before learning the higher level tasks. Algorithms RAR and MAXQ are applied to all the tasks above and including \emph{Navigate}, which also uses the Active Path learning algorithm from section~\ref{sec:avl}. Here is a summary of each task, including its reward model, termination predicate $T_i$, and state abstraction function $Z_i$:

\textbf{Root} - this task selects between \emph{Get} and \emph{Put} to pickup and deliver the passenger. It requires no learning because the termination criteria of the subtasks fully determine when they should be invoked. $T_{Root} = (PassLock =PassDest)$ and $Z_{Root}=\{\}$.

\textbf{Get} - getting the passenger involves navigating through the city, parking the car and picking up the passenger. In this task, the \emph{LegalLoad} state is true when the taxi is at the passenger's location. Receives a reward of $750$ when the passenger is picked up, $T_{Get} = ((PassLoc=0)~or~(PassLoc=PassDest))$, and $Z_{Get}=\{\}$.

\textbf{Put} - similar to \emph{Get}, also receives reward of $750$ when passenger is successfully delivered. The passenger destination \emph{PassDest} is passed to the \emph{Navigate} task. The abstracted \emph{LegalLoad} state is true when the taxi is at the passenger's destination location. $T_{Put} = ((PassLoc>0)~or~(PassLoc=PassDest))$ and $Z_{Put}=\{\}$.

\textbf{Pickup} - this is a primitive action, with a reward of $0$ if successful, and $-2500$ if a pickup is invalid (if the taxi is not stopped, or if \emph{LegalLoad} is false). $Z_{Pickup} = \{LegalLoad, Stopped\}$.

\textbf{Dropoff} - this is a primitive action, with a reward of $1500$ if successful, and $-2500$ if a dropoff is invalid. $Z_{Dropoff} = \{LegalLoad, Stopped\}$.

\textbf{Navigate} - this task learns the sequence of intersections from the current \emph{TaxiLoc} to a target destination \emph{T}. By parameterizing the value function of this task, we can apply Active Path learning as described in Section~\ref{sec:avl}. $T_{Navigate}=(TaxiLoc=\text{\emph{T}})$ and $Z_{Navigate}=\{T, TaxiLoc\}$.

\textbf{Follow} - this is the previously trained continuous trajectory following task that takes as input an adjacent waypoint \emph{WP}, and generates continuous steering and throttle values to follow the straight-line trajectory from \emph{TaxiLoc} to \emph{WP}. $T_{Follow}=(TaxiLoc=\text{\emph{WP}})$ and $Z_{Follow}=\{\text{\emph{WP}}, \Omega_{err}, V_{err}, Y_{err}, V_y\}$.

\textbf{Park} - this is a hard-coded task which simply puts on the brakes ($steer=0$, $throttle=-1$).

\textbf{Drive} - this performs one time step of the physics simulation, with the given steer and throttle inputs. The default reward per time step of driving is $-0.75$.

\subsection{Bayesian Optimization for Hierarchical Policies}
\label{sec:active}

The objective of Bayesian optimization is to learn properties of the value function or policy with as few samples as possible. In direct policy search, where this idea has been explored previously \cite{Martinez--Cantin:2007}, the evaluation of the expected returns using Monte Carlo simulations is very costly. One, therefore, needs to find a peak of this function with as few policy iterations as possible. As shown here, the same problem arises when we want to learn an approximation of the value function only over the relevant regions of the state space. Bayesian optimization provides an exploration-exploitation mechanism for finding these relevant regions and fitting the value function where needed. 

When carrying out direct policy search \cite{Ng:2000}, the Bayesian optimization approach has several advantages over the policy gradients method \cite{Baxter:2001}: it is derivative free, it is less prone to be caught in the first local minimum, and it is explicitly designed to minimize the number of expensive value function evaluations.

\subsubsection{Active Policy Optimization}
\label{sec:apl}

\begin{figure}
    \includegraphics[width=11cm]{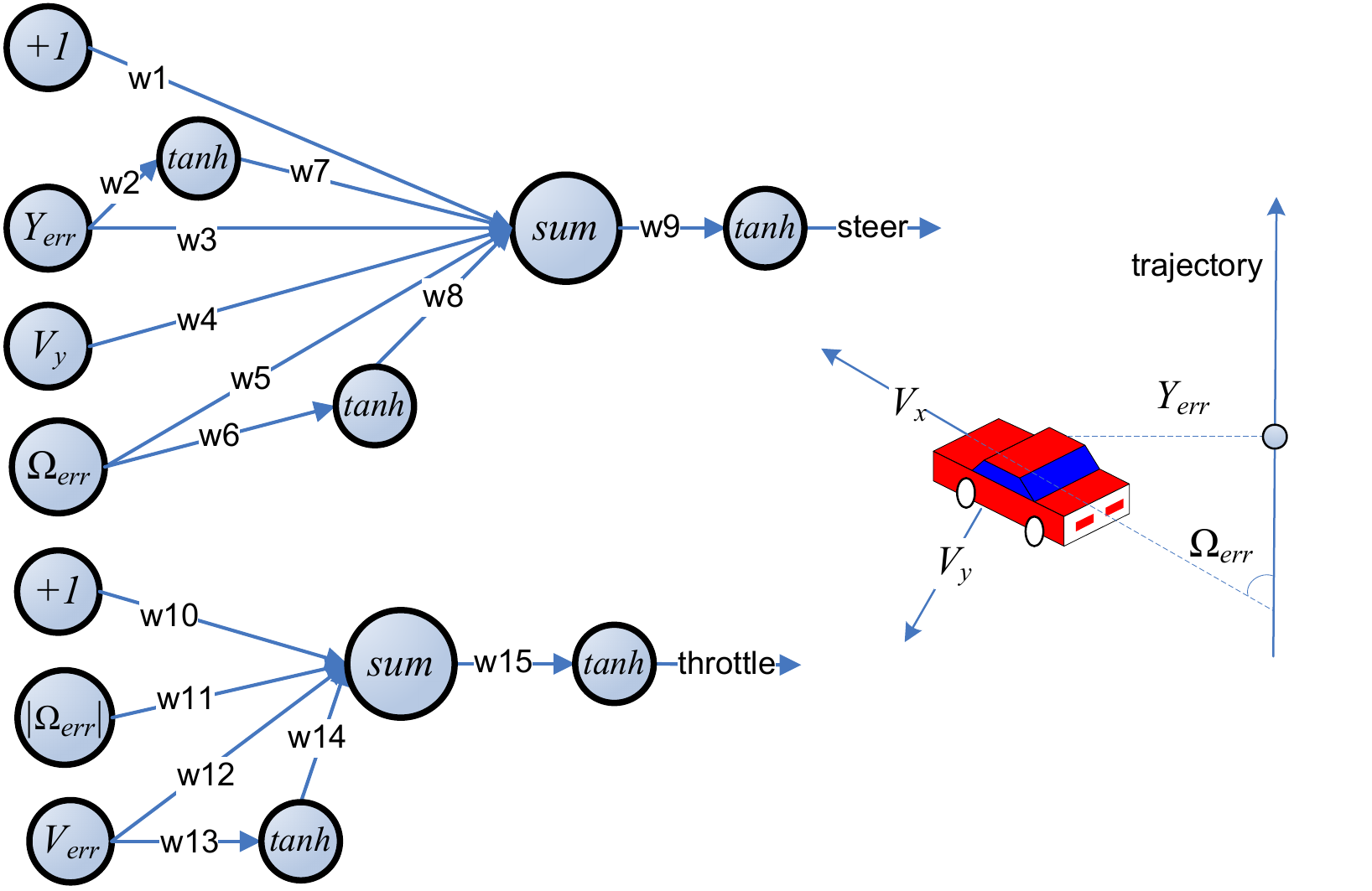}
    \caption{\capstyle{\textbf{Trajectory-following policy:} this parameterized policy, inspired by Ng \emph{et al} [2003] minimizes the error between the vehicle's heading and velocity while following a trajectory. The positional errors  $X_{err}$ and $Y_{err}$ are in trajectory coordinates, $\Omega_{err}$ refers to the difference between the current heading and the trajectory tangent, and $V_{err}$ is the difference between the real and desired velocities.}}
    \label{fig:vehicle_nn}
\end{figure}

\begin{algorithm}
\caption{Bayesian Active Learning with GPs}\label{bal}
\begin{algorithmic}[1]
{\footnotesize
\STATE $N = 0$
\STATE Update the expected improvement function over $D_{1:N}$.
\STATE Choose $\x_{N+1} = \argmax_x EI(\x)$.
\STATE Evaluate $V_{N+1} = V(\x_{N+1})$ and halt if a stopping criterion is met.
\STATE Augment the data $D_{1:N+1} = \{D_{1:N}, (\x_{N+1}, V_{N+1})\}$.
\STATE $N = N+1$ and go to step 2.
}
\end{algorithmic}
\end{algorithm}

The lowest level \emph{Drive} task uses the parameterized function illustrated in Figure \ref{fig:vehicle_nn} to generate continuous steer and throttle values, within the range of {-1 to 1}. The $|\x|=15$ parameters (weights) are trained using the Bayesian active policy learning Algorithm \ref{bal}. We first generate and evaluate a set of $30$ Latin hypercube samples of $\x$ and store them and corresponding values vector $V$ in the data matrix $\data$. The value of a trajectory is the negative accumulated error between the car's position and velocity, and the desired position and velocity. The policy evaluation consists of averaging $10$ episodes along the same trajectory but with different, evenly spaced starting angles, where the car needs to accelerate from rest, go to the first waypoint, perform a u-turn, and arrive back to the starting location. In a noisier environment, more samples would be necessary to properly evaluate a policy. The TORCS simulator is deterministic, and a small amount of noise arises from unmodeled tire slipping and random bumpiness of the road. The $10$ different starting angles were sufficient for evaluating a policy in our experiments. Subsequently, we perform the iteration described in Algorithm~\ref{bal} to search for the best instantiation of the parameters.

\subsubsection{Active Value Function Learning}
\label{sec:avl}

The \emph{Navigate} task learns path finding from any intersection in the topological map to any of the destinations. Although this task operates on a discrete set of waypoints, the underlying map coordinates are continuous, and we can again apply active exploration with GPs.

Unlike the previous algorithm that searches for a set of optimal parameters, Algorithm~\ref{avl} learns the value function at a finite set of states, by actively generating exploratory actions; it is designed to fit within a MAXQ task hierarchy. The 4-dimensional value function $V(\x)$ in this case is parameterized by two 2D map coordinates $\x=\{x_C, y_C, x_T, y_T\}$, and stores the sum of discounted rewards while travelling from the current intersection $|C|=61$ to the target $|T|=22$. The 1342 sampled instances of $\x_{1:1342}$ and corresponding $V$ vector are stored in the data matrix $\data$; it is initialized with $V(x_T,y_T,x_T,y_T)=0$ for all target destinations $T$, which enables the GP to create a useful response surface without actually having observed anything yet.

In the $\epsilon-$greedy experiments, a random intersection is chosen with chance $0.1$, and the greedy one with chance $0.9$. For the active exploration case, we fit a GP over the data matrix $\data$, and pick the adjacent intersection that maximizes the expected improvement function. We parameterize this function with an annealing parameter that decays over time such that initially we place more importance on exploring.

\begin{algorithm}[t!]
\caption{Active Path Learning with GPs} \label{avl}
\begin{algorithmic}[1]
{
\STATE {\bf function} $Navigate Task Learner( Navigate~i, State~s )$\\[5pt]
   \STATE {\bf let} $trajectory$=() - list of all states visited in $i$
   \STATE {\bf let} $intersections$=() - intersection states visited in $i$

   \STATE {\bf let} $visits=0$ - \# of visits at an intersection in $i$ \\[5pt]
   \WHILE{$Terminated_i(s)$ is false}

       \STATE choose adjacent intersection \emph{WP} using $\epsilon$-greedy or Active exploration.
       \STATE {\bf let} $childSeq = Follow(\text{\emph{WP}},s)$
       \STATE {\bf append} $childSeq$ onto the front of $trajectory$
       \STATE observe result state $s'$ \\[5pt]

       \STATE $N=$ length( $childSeq$ )
       \STATE $R = \sum_{j=1}^{N} \gamma^{N-j}\times r_j$ be the total discounted reward received from $s$ to $s'$

        \STATE $V_s' = V( TaxiLoc_{s'}, Target_i )$ \COMMENT{ guaranteed $<=0$}
        \STATE $V_s = V( TaxiLoc_s, Target_i )$ \COMMENT{ guaranteed $<=0$}

       \IF{ $Terminated_i(s')$ is true }
           \STATE $V_s \leftarrow (1-\alpha) \times V_s + \alpha \times R$
           \FORALL{ $j = 1~\text{to length(intersections)}$ }
               \STATE $\{s', N', R'\}$ = intersections($j$)
               \STATE $R \leftarrow R' + \gamma^{N'} \times R$
               \STATE $V_s' \leftarrow V( TaxiLoc_{s'}, Target_i )$
               \STATE $V_s' \leftarrow (1-\alpha) \times V_s' + \alpha \times R$
           \ENDFOR \\[5pt]
       \ELSE
           \STATE {\bf append} $\{s, N, R\}$ onto the front of $intersections$

           \STATE $visits(TaxiLoc_s) \leftarrow visits(TaxiLoc_s) + 1$

           \STATE $penalty \leftarrow V_s \times visits(TaxiLoc_s)$ \COMMENT{prevent loops}
           \STATE $V_s \leftarrow (1-\alpha) \times V_s + \alpha ( penalty + R + \gamma^N \times  V_s')$
       \ENDIF \\[5pt]

       \STATE $s = s'$
   \ENDWHILE \\[5pt]

   \RETURN{ $trajectory$ }
}
\end{algorithmic}
\end{algorithm}

The true value will not be known until the $Navigate$ task reaches its destination and terminates, but we still need to mark visited intersections to avoid indefinite looping. Lines 23--26 compute an estimated value for $V(s)$ by summing the immediate discounted reward of executing $Follow(\text{\emph{WP}},s)$ with the discounted, previously recorded value of the new state $V(s')$, and a heuristic penalty factor to avoid looping. Once we reach the destination of this task, $Target_i$, we have the necessary information to propagate the discounted reward to all the intersections along the trajectory, in lines 15--21.


\afterpage{\clearpage}

 \begin{figure}[f]
     \centering
    \includegraphics[width=10cm]{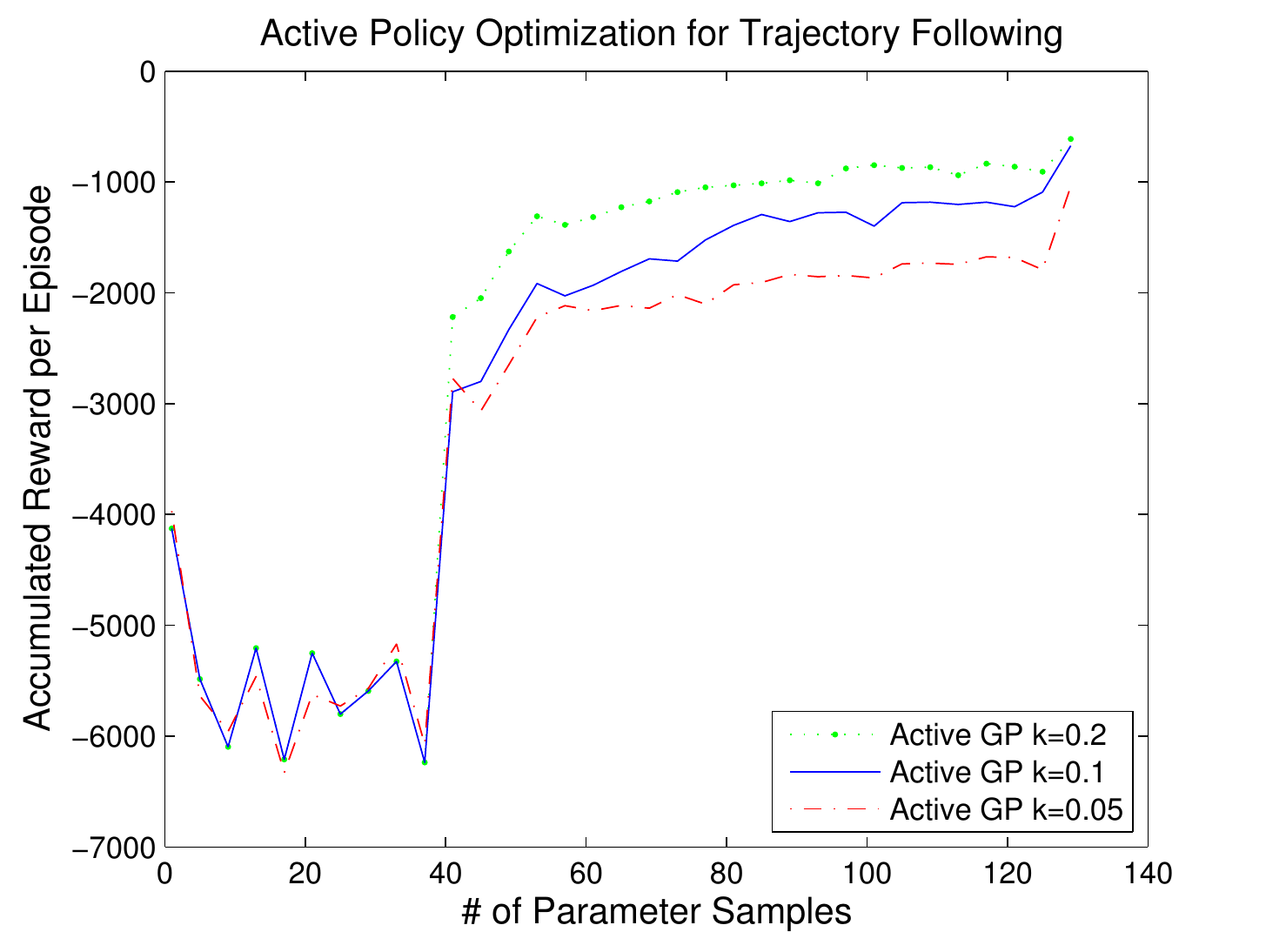}\vspace{-0.5cm}
    \caption{\capstyle{\textbf{Active Policy Optimizer:} searching for the $15$ policy parameters, and comparing different values for the GP kernel size $k$. We used the Expected Improvement function \ref{eqn:EGO_EI}, and the three experiments are initialized with the the same set of $30$ Latin hypercube samples. A total of $20$ experimental runs were averaged for this plot.}}\label{fig:res_traj_kernel}
\end{figure}

\begin{figure}[f]
    \centering
    \includegraphics[width=10cm]{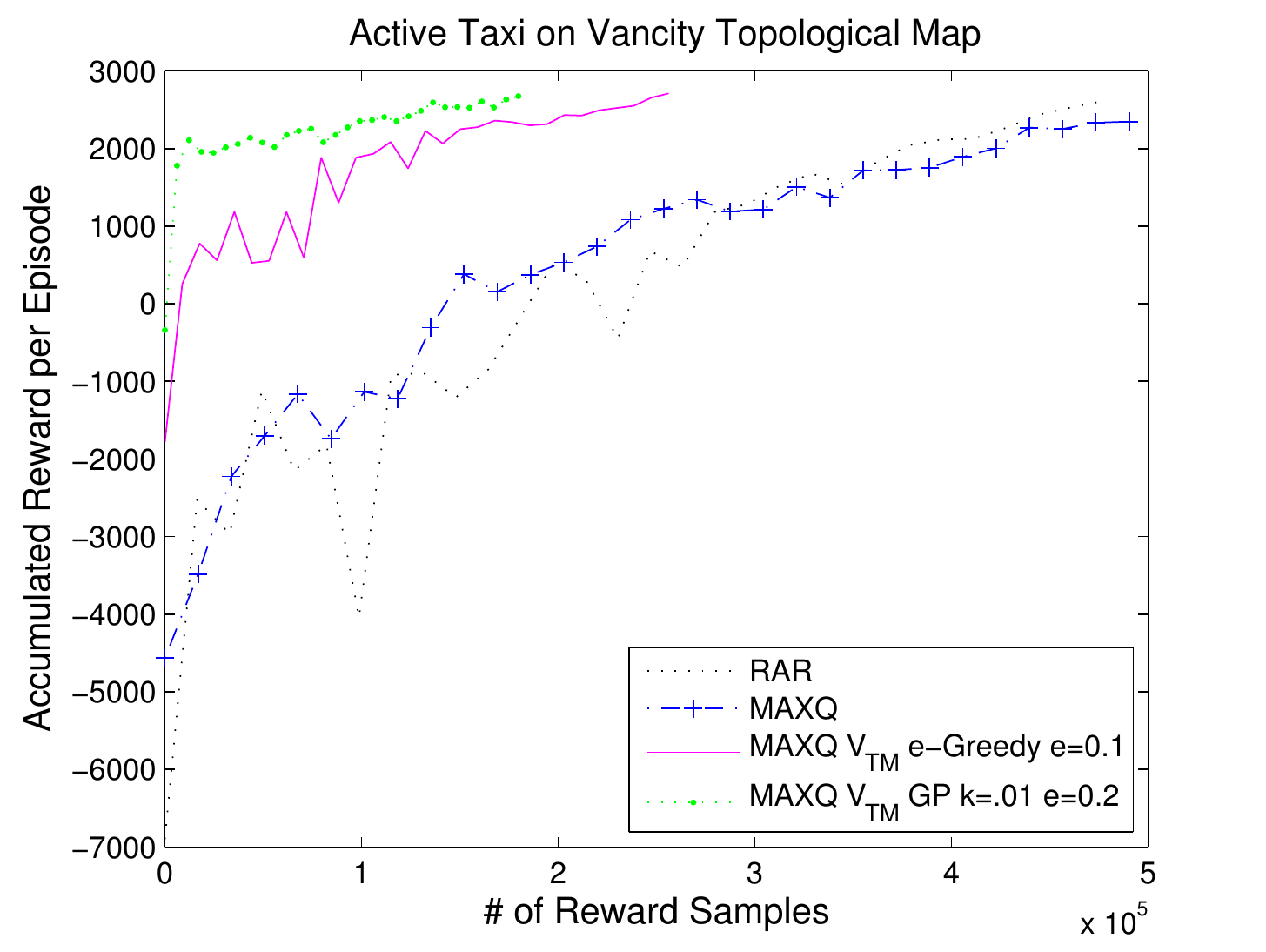}
    \caption{\capstyle{\textbf{Parameterized $V_{TM}$ vs. RAR and MAXQ:} These experiments compare the original Recursive Average Reward (RAR) and MAXQ (discounted reward) algorithms against the parameterized $V_{TM}(TaxiLoc,\text{\emph{WP}})$ path learner. }}
    \label{fig:res_vancity_rar}
\end{figure}

\begin{figure}[h!]
   \centering
   \subfigure[$V_{TM}~GP~k=0.01$] {
       \includegraphics[width=10cm]{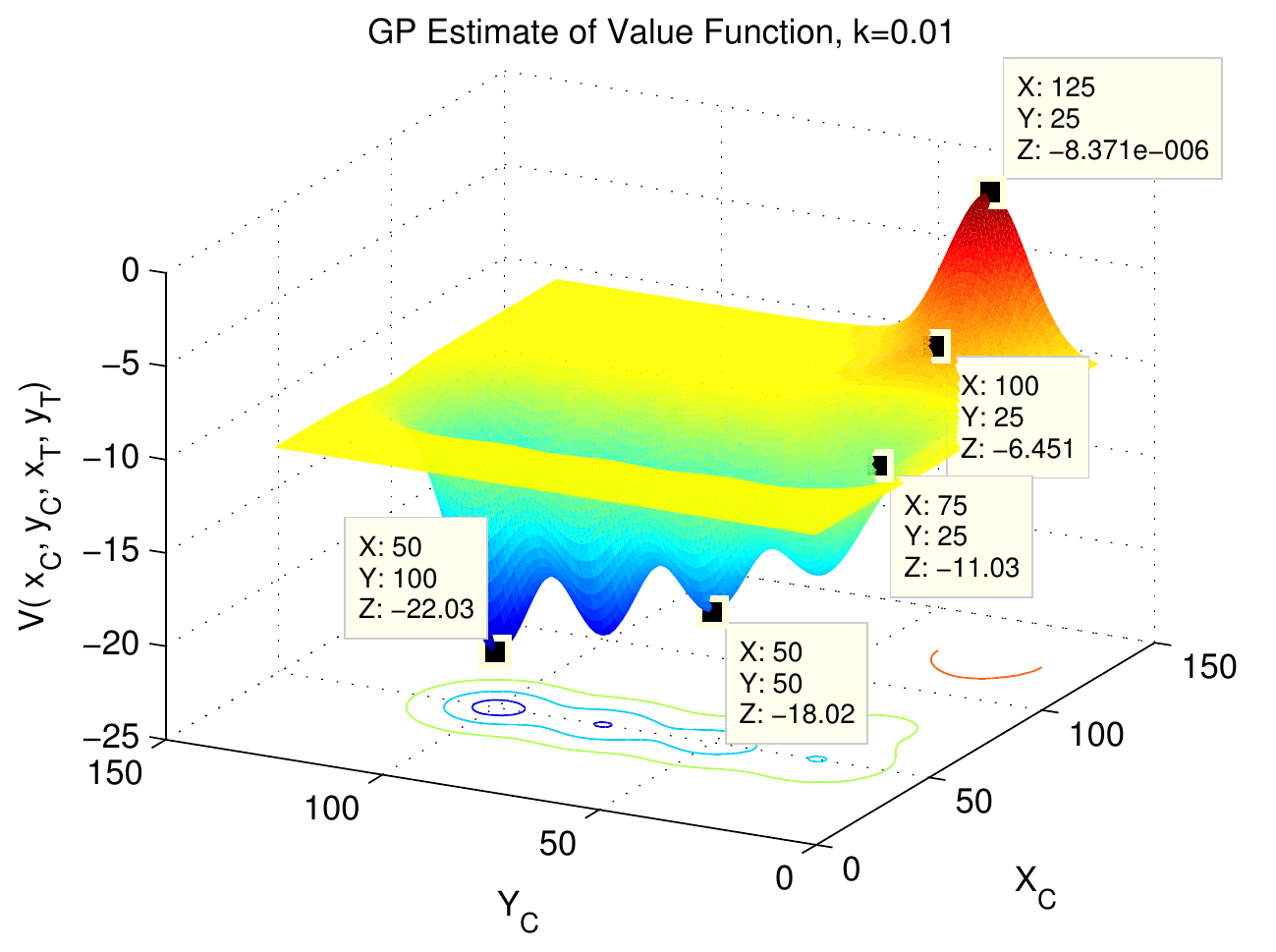}}
   \subfigure[$V_{TM}~GP~k=0.02$] {
       \includegraphics[width=13cm]{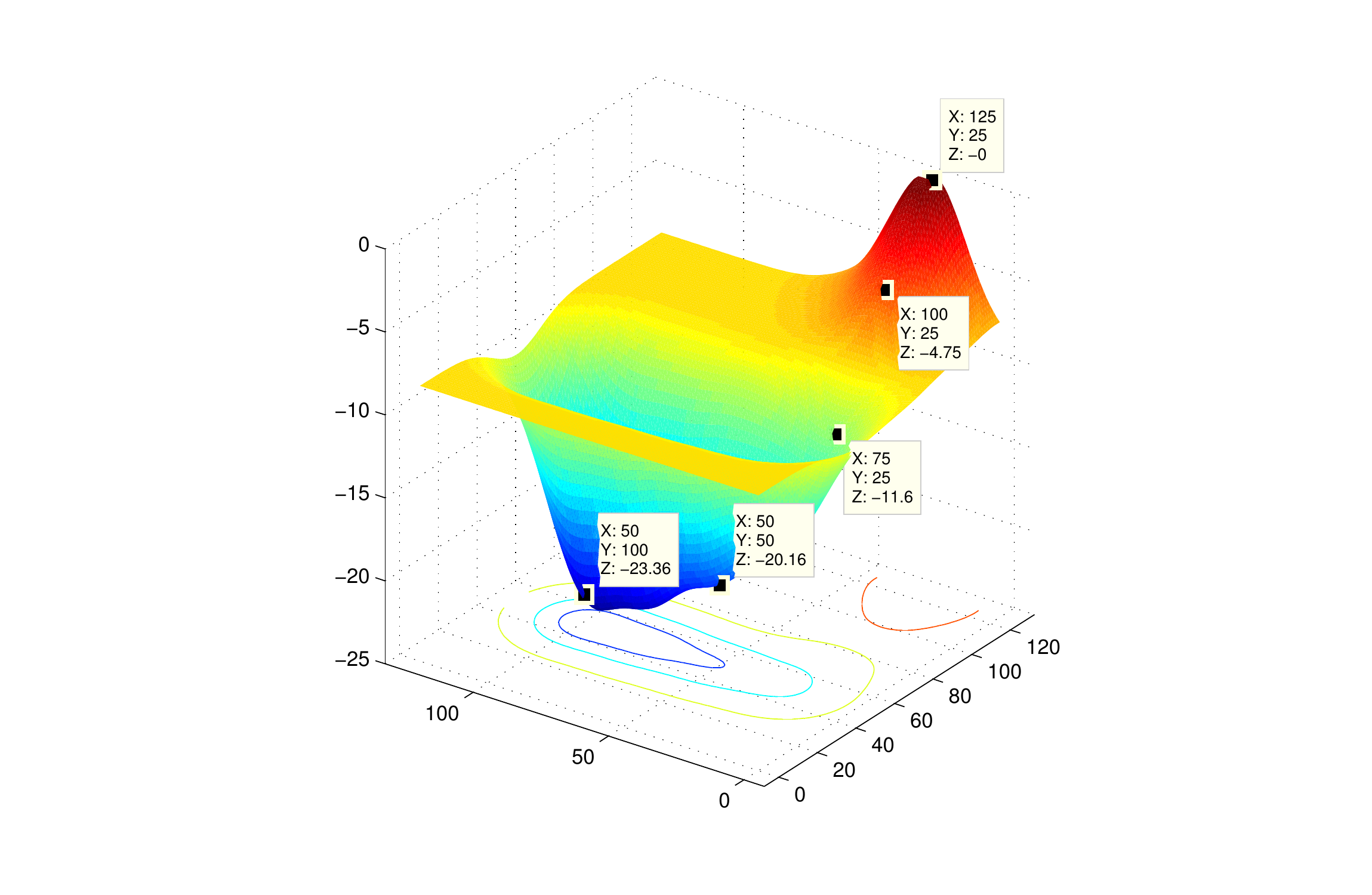}}
   \caption{\capstyle{\textbf{GP Response Surface.} A small kernel value narrows the `footprint' of an observation, whereas a larger $k$ interpolates to the surrounding state space.}}
   \label{fig:gp_k02}
\end{figure}

\subsection{Simulations}
\label{sec:experiments}

The nature of the domain requires that we run policy optimization first to train the \emph{Follow} task. This is reasonable, since the agent cannot be expected to learn map navigation before learning to drive the car. Figure~\ref{fig:res_traj_kernel} compares the results of three different values for the GP kernel $k$, when running the active policy optimization algorithm from \Section \ref{sec:apl}. The desired velocity is $60km/hr$, a time step lasts $0.25$~seconds, and the trajectory reward $R= - \sum_t{ \left[ 1 \times \tilde Y_{err}^2 + 0.8 \times \tilde V_{err}^2 + 1 \times \tilde \Omega_{err}^2 + 0.7 \times \bf{\tilde a' \tilde a} \right]}$ is the negative weighted sum of normalized squared error values between the vehicle and the desired trajectory, including $\bf{a}=\left[steer,throttle\right]$ to penalize for abrupt actions. After $\sim50$ more parameter samples (after the initial $30$ random samples), the learner has already found a useable policy.

Subsequently, the best parameters are fixed inside the \emph{Follow} task, and we run the full hierarchical learners, with results in Figure~\ref{fig:res_vancity_rar}. We averaged the results from 10 runs of RAR, MAXQ, and the value learning Algorithm~\ref{avl} applied only to the \emph{Navigate} task (with the rest of the hierarchy using MAXQ). All the experiments use the hierarchical task model presented in Section~\ref{sec:hierarchy}. Each reward time step lasts $0.3$ seconds, so the fastest learner, $V_{TM}~GP$ with $\epsilon=0.2$ drove for $\sim4$ hours real-time at $\sim60~km/hr$ before finding a good approximation of the $V_{Navigate}$ value function. Refer to Figure~\ref{fig:gp_k02} for an intuition of how fitting the GP over the samples values transfers observations to adjacent areas of the state space. This effect is controlled through the GP kernel parameter $k$. While the application is specific to navigating a topological map, the algorithm is general and can be applied to any continuous state spaces of reasonable dimensionality.

\section{Discussion and advice to practitioners}\label{sec:discussion}

Bayesian optimization is a powerful tool for machine learning, where the problem is often not acquiring data, but acquiring labels.  In many ways, it is like conventional active learning, but instead of acquiring training data for classification or regression, it allows us to develop frameworks to efficiently solve novel kinds of learning problems such as those discussed in Sections~~\ref{sec:prefgalleries} and \ref{sec:hiercontrol}.  It proves us with an efficient way to learn the solutions to problems, and to collect data, all within a Bayesian framework.

However, Bayesian optimization is also a fairly recent addition to the machine learning community, and not yet extensively studied on user applications.  Here, we wish to describe some of the shortcomings we have experienced in our work with Bayesian optimization, both as caveats and as opportunities for other researchers.  

A particular issue is that the design of the prior is absolutely critical to efficient Bayesian optimization.  Gaussian processes are not always the best or easiest solution, but even when they are, great care must be taken in the design of the kernel.  In many cases, though, little is known about the objective function, and, of course, it is expensive to sample from (or we wouldn't need to use Bayesian optimization in the first place).  The practical result is that in the absence of (expensive) data, either strong assumptions are made without certainty that they hold, or a weak prior must be used.  It is also often unclear how to handle the trade-off between exploration and exploitation in the acquisition function.  Too much exploration, and many iterations can go by without improvement.  Too much exploitation leads to local maximization.  

These problems are exacerbated as dimensionality is increased---more dimensions means more samples are required to cover the space, and more parameters and hyperparameters may need to be tuned, as well.  In order to deal with this problem effectively, it may be necessary to do automatic feature selection, or assume independence and optimize each dimension individually.


Another limitation of Bayesian optimization is that the acquisition is currently both myopic and permits only a single sample per iteration.  Looking forward to some horizon would be extremely valuable for reinforcement learning problems, as well as in trying to optimize within a known budget of future observations. Recent work \cite{Garnett:2010a,Azimi:2011} has indicated very promising directions for this work to follow.  Being able to efficiently select entire sets of samples to be labelled at each iteration would be a boon to design galleries and other batch-incremental problems.

Finally, there are many extensions that will need to be made to Bayesian optimization for particular applications---feature selection, time-varying models, censored data, heteroskedasticity, nonstationarity, non-Gaussian noise, \emph{etc}.  In many cases, these can be dealt with as extensions to the prior---in the case of Gaussian processes, for example, a rich body of literature exists in which such extensions have been proposed.  However, these extensions need to take into account the adaptive and iterative nature of the optimization problem, which can vary from trivial to impossible.  

Clearly, there is a lot of work to be done in Bayesian optimization, but we feel that the doors it opens make it worthwhile.  It is our hope that as Bayesian optimization proves itself useful in the machine learning domain, the community will embrace the fascinating new problems and applications it opens up.

\bibliographystyle{named}
\begin{small}
\bibliography{thesis}
\end{small}
\end{document}